\documentclass[10pt,twocolumn,letterpaper]{article}

\usepackage[pagenumbers]{cvpr} 

\usepackage{tablefootnote}
\usepackage[dvipsnames]{xcolor}

\usepackage{graphicx}
\usepackage{amsmath}
\usepackage{amssymb}
\usepackage{booktabs}
\usepackage{multirow}
\usepackage{dblfloatfix}

\definecolor{cvprblue}{rgb}{0.21,0.49,0.74}
\usepackage[pagebackref,breaklinks,colorlinks,citecolor=cvprblue]{hyperref}

\usepackage[capitalize]{cleveref}
\crefname{section}{Sec.}{Secs.}
\Crefname{section}{Section}{Sections}
\Crefname{table}{Table}{Tables}
\crefname{table}{Tab.}{Tabs.}

\def\Vec#1{{\boldsymbol{#1}}}
\def\Mat#1{{\boldsymbol{#1}}}

\def\paperID{532} 
\def\confName{CVPR}
\def\confYear{2024}

\title{Beyond Average: 
Individualized Visual Scanpath Prediction}

\author{Xianyu Chen\qquad Ming Jiang\qquad Qi Zhao\\
University of Minnesota, United States\\
{\tt\small \{chen6582, mjiang\}@umn.edu, qzhao@cs.umn.edu}
}

\begin{document}
\maketitle
\begin{abstract}
Understanding how attention varies across individuals has significant scientific and societal impacts. However, existing visual scanpath models treat attention uniformly, neglecting individual differences. To bridge this gap, this paper focuses on individualized scanpath prediction (ISP), a new attention modeling task that aims to accurately predict how different individuals shift their attention in diverse visual tasks. It proposes an ISP method featuring three novel technical components: (1) an observer encoder to characterize and integrate an observer's unique attention traits, (2) an observer-centric feature integration approach that holistically combines visual features, task guidance, and observer-specific characteristics, and (3) an adaptive fixation prioritization mechanism that refines scanpath predictions by dynamically prioritizing semantic feature maps based on individual observers' attention traits. These novel components allow scanpath models to effectively address the attention variations across different observers. Our method is generally applicable to different datasets, model architectures, and visual tasks, offering a comprehensive tool for transforming general scanpath models into individualized ones. Comprehensive evaluations using value-based and ranking-based metrics verify the method's effectiveness and generalizability. 
\end{abstract}

\section{Introduction}
\label{sec:intro}

\begin{figure}[t]
\centering
\includegraphics[width=1.0\linewidth]{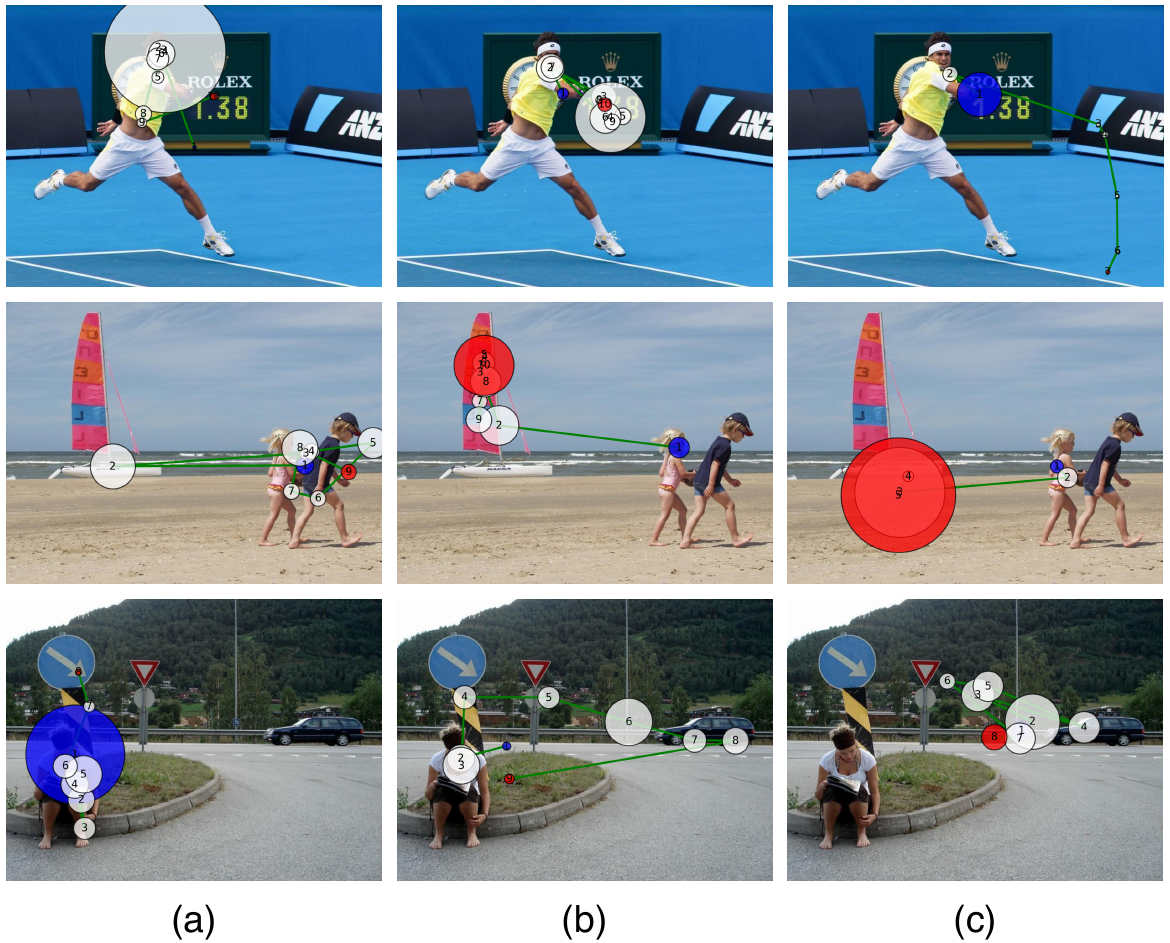}
\caption{Understanding and predicting the distinct eye movements of each observer is the key objective of individualized scanpath prediction. These examples reveal the variations in the scanpaths of different observers, showing their distinct attention preferences in (a) faces, (b) objects, and (c) background. Each dot represents a fixation, with the number and radius indicating its order and duration, respectively. The \textcolor{blue}{blue} and \textcolor{red}{red} dots indicate the beginning and the end of the scanpath, respectively.}
\label{fig:teaser}
\end{figure}


Saccadic eye movements, such as fixations and saccades, enable individuals to shift their attention quickly and redirect their focus to different points in the visual field. Studying various factors driving people's eye movements is important for understanding human attention and developing human-like attention systems.
Computational models predicting eye movements have broad impacts across various domains, such as assessing image and video quality~\cite{ke:2016:saliencyquality,patrick:2013:attentionapplication,leida:2016:qualityassessment}, developing intuitive human-computer interaction systems~\cite{tommy:2009:eyetrackinghci,thomas:1989:hcieyegaze,uchenna:2021:eyegazehci,anjana:2013:hcieyegaze,yue:2023:ueyes}, creating immersive virtual reality experiences~\cite{thammathip:2017:eyegazevr,isayas:2023:hcieyegaze,kun:2021:hcieyegaze}, enhancing the safety and efficiency of autonomous vehicles~\cite{xinyue:2022:gazeselfdriving,ye:2018:gazeselfdriving,ye:2019:gazeguideselfdriving}, and diagnosing neurodevelopmental conditions~\cite{ming:2020:asd,shi:2019:asd,huiyu:2019:saliency4asd}.

While existing models of saccadic eye movements predominantly focus on modeling generic gaze patterns manifested as observer-agnostic scanpaths (\ie, a spatiotemporal sequence of fixations), this work seeks to model the individual variations in eye movements. As shown in Figure~\ref{fig:teaser}, there exists significant inter-observer variations in visual scanpaths. Such variations can be attributed to a multitude of individual characteristics, such as gender, age, and neurodevelopmental conditions~\cite{evan:2012:curiouseye,matthew:2013:individualdifference}.
For instance, females show more explorative gaze patterns than males~\cite{bahman:2019:gendergazeindoor,felix:2012:gendergazeface,negar:2017:gendergazeface}, older adults prefer faces~\cite{young:2020:agegaze} and objects with high color visibility~\cite{zeyu:2022:agegaze}, individuals with neurodevelopmental disorders, such as autism spectrum disorder (ASD), may show a preference for repetitive patterns while avoiding social cues~\cite{shuo:2015:austim,mikle:2005:austim,mark:1998:austim}. Therefore, developing tailored models that cater to the uniqueness of each observer is an essential step toward more precise and adaptive attention modeling.

Existing research efforts have failed to address the divergence between the personalized nature of human attention and the collective nature of current scanpath models. This is due to the lack of standardized methods for quantifying and representing individual attention traits, as well as the absence of comprehensive frameworks that can accommodate the diverse range of observer characteristics.  In this paper, we resolve this significant challenge with a novel individualized scanpath prediction (ISP) method comprising three novel components: (1) The observer encoder is a key component for personalized scanpath modeling. It efficiently captures an observer's unique attention traits by introducing an observer-specific identifier as an additional input, forming the basis for individualized scanpath predictions.
(2) The observer-centric feature integration module adopts a comprehensive approach, fusing visual features, task guidance, and observer-specific attention traits spatially and channel-wise. This ensures consideration of diverse bottom-up and top-down cues, simplifying subsequent processing and enhancing the efficient prediction of individualized scanpaths.
(3) The adaptive fixation prioritization module enhances scanpath precision by dynamically assigning priorities to the output features, generating a probability map for each fixation. This adaptability ensures refined predictions of individualized scanpaths.

Our method has three distinctions from previous visual scanpath studies: (1) We go beyond prior work focusing on general scanpath modeling and propose the first comprehensive investigation of individualized scanpath prediction. (2) We emphasize the tight integration of observer features into the scanpath prediction process, distinct from trivial individualization techniques such as fine-tuning with single-observer data. (3) Our method is generally applicable to various model architectures and visual tasks, broadening its usability in real-world applications.

The main contributions of this work are as follows:

\begin{enumerate}
    \item We study the underexplored task of individualized scanpath prediction, focusing on modeling how an observer's unique attention traits affect their eye movements. 
    \item 
    We propose an individualization method featuring three novel technical components: The observer encoder is an important addition to scanpath models, which enables observer-centric feature integration and adaptive fixation prioritization. These components enable the model to adapt to individual observers, yielding accurate and individualized predictions.
    \item 
    We comprehensively evaluate scanpaths from individual observers' perspectives, using both value-based and ranking-based metrics. Experimental results on multiple eye-tracking datasets, with different model architectures and visual tasks, prove our method's effectiveness and generalizability for predicting individualized scanpaths.
\end{enumerate}

\begin{figure*}[t]
\centering
\includegraphics[width=1.0\linewidth]{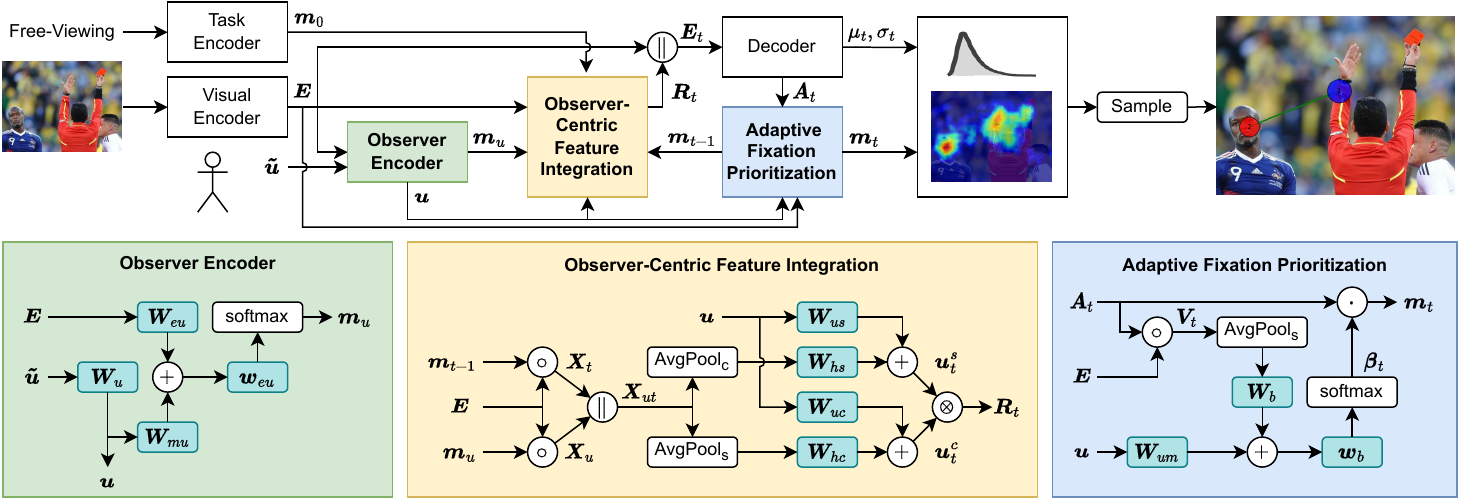}
\caption{Our proposed method incorporates an observer encoder for characterizing individualized attention traits, followed by observer-centric feature integration for holistic processing, and adaptive fixation prioritization for refined predictions.}
\label{fig:architecture}
\end{figure*}

\section{Related Works}
\label{sec:related-work}
Our work is related to prior studies on eye-tracking datasets and visual scanpath prediction methods.

\subsection{Eye-Tracking Datasets}
The foundation for attention modeling relies on diverse, thoughtfully curated eye-tracking datasets spanning various stimuli, tasks, and observers~\cite{shi:2020:air,shuo:2015:austim,juan:2014:osie,zhibo:2020:cocosearch,huiyu:2019:saliency4asd}. These datasets, from those dedicated to free-viewing~\cite{shuo:2015:austim,juan:2014:osie,huiyu:2019:saliency4asd} to those capturing goal-directed behaviors~\cite{shi:2020:air,zhibo:2020:cocosearch}, serve as invaluable resources for training and evaluating attention models. Specifically, several well-recognized eye-tracking datasets have provided benchmarks to quantify the performance of saliency models~\cite{ming:2015:salicon,tilke:2013:mit1003,juan:2014:osie,ali:2015:cat2000} and scanpath models~\cite{shuo:2015:austim,juan:2014:osie,huiyu:2019:saliency4asd}. Subsequent studies have developed datasets of goal-directed behaviors to characterize how observers search for an object in an image~\cite{zhibo:2020:cocosearch} or answer image-related questions~\cite{shi:2020:air}. These efforts facilitate the development of static saliency models~\cite{matthias:2016:deepgaze,marcella:2018:sam,xun:2015:salicon,camilo:2020:umsi,souradeep:2022:agdf,sen:2020:eml,shi:2023:personalsaliency,bahar:2023:tempsal} as well as dynamic scanpath models~\cite{xianyu:2021:vqa,sounak:2023:gazeformer,wanjie:2019:iorroi,zhibo:2020:cocosearch,zhibo:2023:humanattention,zhibo:2022:targetabsent,ryan:2022:scanpathnet,mengyu:2023:scanpath,xiangjie:2023:scandmm}. Our work sets itself apart from individualized saliency models~\cite{yanyu:2018:personalsaliency,yanyu:2017:personalsaliency,yuya:2020:fewshotpersonalsaliency,shi:2023:personalsaliency,anqi:2017:individualscanpath,xinhui:2022:fewshotsaliency} by predicting dynamic scanpaths rather than static saliency maps. It utilizes datasets from various visual tasks and observer groups to expand scanpath modeling, with emphasis on the distinct attention traits of each observer.

\subsection{Visual Scanpath Prediction}
Scanpath prediction has been an underexplored topic in the field of attention modeling. Early studies generate scanpaths by sampling fixations from saliency maps using the inhibition-of-return mechanism~\cite{wei:2011:stde,olivier:2015:saccadicmodel,calden:2018:star-fc,laurent:1998:visualattention,yixiu:2017:scanpathestimation,anqi:2017:individualscanpath}. Recent studies have developed computational models directly predicting the sequence of fixations and saccades~\cite{xianyu:2021:vqa,sounak:2023:gazeformer,wanjie:2019:iorroi,zhibo:2020:cocosearch,zhibo:2023:humanattention,zhibo:2022:targetabsent,ryan:2022:scanpathnet,mengyu:2023:scanpath}. Several scanpath models harness the power of deep neural networks~\cite{xianyu:2021:vqa,sounak:2023:gazeformer,wanjie:2019:iorroi,zhibo:2020:cocosearch,zhibo:2023:humanattention,zhibo:2022:targetabsent,ryan:2022:scanpathnet,mengyu:2023:scanpath,matthias:2022:deepgaze,yue:2023:ueyes}, reinforcement learning techniques~\cite{xianyu:2021:vqa,zhibo:2020:cocosearch,zhibo:2022:targetabsent}, and transformer-based models~\cite{mengyu:2023:scanpath,sounak:2023:gazeformer}, ultimately improving the accuracy of scanpath prediction to the human level. These developments have significantly deepened our understanding of the temporal dynamics of human attention. However, existing models focus on predicting general scanpaths rather than taking individual variations into account. Differently, our method places particular emphasis on characterizing individual attention traits and integrating them into a general scanpath model, thus enabling tailored predictions that align with each observer's gaze behavior. This unique approach extends the horizon of attention modeling, underlining the importance of individual differences within the broader context of human attention.

\section{Methodology}
\label{sec:methodology}

The core challenge in individualized scanpath modeling is the need to predict unique gaze patterns for different observers. This arises due to the inherent variations in attention traits. Figure~\ref{fig:architecture} presents an overview of our method. It offers a threefold solution: (1) an observer encoder, (2) an observer-centric feature integration module, and (3) an adaptive fixation prioritization module. 
These components are designed to be flexible as they can be applied on a general scanpath model based on the encoder-decoders, (\eg, with a visual encoder and task encoder, and an LSTM~\cite{sepp:1997:lstm} or Transformer~\cite{jacob:2018:bert} decoder) to provide robust and precise predictions tailored to each observer.

\subsection{Observer Encoding}

At the core of our proposed method is an \textit{Observer Encoder}, a key component designed to enable the novel task of individualized scanpath prediction. It takes as input an observer-specific identifier $\Tilde{\Vec{u}}$ (\eg, a one-hot vector) and efficiently computes an observer feature $\Vec{u}$. This feature represents the unique characteristics and preferences of each observer. Our approach utilizes a linear embedding operation to derive the observer feature:
\begin{equation}
    \Vec{u} = \Vec{W}_{u} \Tilde{\Vec{u}},
    \label{equ:user-vector}
\end{equation}
where $\Vec{W}_{u}$ indicates learnable parameters. The linear embedding operation provides a straightforward mapping that retains important characteristics, offering a practical and computationally efficient solution for capturing unique attention traits.

This observer encoder can be seamlessly integrated into an existing scanpath model. As shown in Figure~\ref{fig:architecture}, a typical \textit{Visual Encoder} is used to transform the input image into multi-channel feature maps $\Vec{E}$ characterizing the bottom-up attention. To model the interaction between the visual feature  $\Vec{E}$ and the observer feature $\Vec{u}$, an observer guidance map can be computed through a linear combination:
\begin{equation}
    \Vec{m}_u = {\rm{softmax}}\big(\Vec{w}_{eu}^T \tanh (\Mat{W}_{eu} \Vec{E} + \Mat{W}_{mu} \Vec{u} )\big),
    \label{equ:attend-map-v-u}
\end{equation}
where $\Vec{w}_{eu}$, $\Mat{W}_{eu}$, $\Mat{W}_{mu}$ are learnable parameters. This observer guidance localizes salient image regions of specific interest to the observer.

Some scanpath models use a \textit{Task Encoder} to process task-relevant information guiding the gaze behavior, such as a search target or a general question to answer. Such top-down guidance can be represented as a spatial attention map $\Vec{m}_0$ prioritizing task-relevant regions. These bottom-up and top-down features are typically processed with a decoder (\eg, LSTM or Transformer) to predict a sequence of probability maps $\{\Vec{m}_t \, | \, t=1,2,\ldots,T\}$ and distribution parameters $\{(\mu_t, \sigma_t^2)\, | \, t=1,2,\ldots,T\}$  for sampling fixation positions and durations, respectively, where $T$ is the number of fixations.

In Sections~\ref{sec:method1} and~\ref{sec:method2}, we present specific modules that leverage the observer feature $\Vec{u}$ to individualize the scanpath model. For clarity, our method description focuses on its integration with an LSTM model~\cite{xianyu:2021:vqa}. Please refer to Section~\ref{sec:implementation} and Supplementary Material for details about its adaptation to a Transformer network~\cite{sounak:2023:gazeformer}.

\subsection{Observer-Centric Feature Integration} 
\label{sec:method1}

With the encoded observer features characterizing each observer's distinct attention traits, we design observer-centric feature integration to address the critical need to fuse various inputs, including visual features, task relevance, and observer-specific characteristics, into a unified representation. The motivation behind this integration is to create a comprehensive understanding of individualized attention patterns. This integration process results in a sequence of observer-centric feature maps $\{\Vec{R}_t \, | \, t=1,2,\ldots,T\}$ representing spatiotemporal fixation patterns, thus enabling the model to track individualized attention dynamics over time~\cite{xianyu:2021:vqa,shi:2020:air,ming:2020:ivqa}.

Specifically, to guide the prediction at each step, we leverage the predicted fixation distribution from the previous step (\ie, $\Vec{m}_{t-1}$) as a soft attention map, applying it to the visual features to derive the previously fixated visual features $\Vec{X}_t = \Vec{E} \circ \Vec{m}_{t-1}$, where the symbol $\circ$ denotes the Hadamard product. It is noteworthy that the task guidance map $\Vec{m}_0$ is used initially to guide the first fixation, mimicking the cognitive process that initially directs eye movements based on the visual task. 
Similarly, the observer guidance map $\Vec{m}_u$ is used as the attention weights to obtain observer-centric visual features $\Vec{X}_u = \Vec{E} \circ \Vec{m}_u$.

To seamlessly integrate the fixated visual features and observer-centric visual features, we concatenate the two types of feature maps
\begin{equation}
    \Mat{X}_{ut} = \Vec{X}_t\mathbin\Vert \Vec{X}_u,
    \label{equ:attend-map-v-ut}
\end{equation}
and perform spatial and channel-wise feature fusion, which are achieved by average-pooling the feature maps along the channel ($\text{AvgPool}_{c}$) and spatial dimensions ($\text{AvgPool}_{s}$), respectively, followed by linear layer processing and the addition of encoded observer features:
\begin{align}
    \Vec{u}^s_t &= \text{ReLU}(\Vec{W}_{hs} \text{AvgPool}_{c}(\Mat{X}_{ut}) + \Vec{b}_{hs}) + \Vec{W}_{us}\Vec{u}, \label{equ:ust-vector}\\
\Vec{u}^c_t &= \text{ReLU}(\Vec{W}_{hc} \text{AvgPool}_{s}(\Mat{X}_{ut}) + \Vec{b}_{hc}) + \Vec{W}_{uc} \Vec{u},
\label{equ:uct-vector}
\end{align}
where $\Vec{W}_{hs}$, $\Vec{W}_{hc}$, $\Vec{W}_{us}$, $\Vec{W}_{uc}$, $\Vec{b}_{hs}$, and $\Vec{b}_{hc}$ are learnable parameters.
Ultimately, combining $\Vec{u}^s_t$ and $\Vec{u}^c_t$ yields the final observer-centric feature maps 
\begin{align}
    \Vec{R}_t=\Vec{u}^s_t \otimes \Vec{u}^c_t,
    \label{equ:recalled-features}
\end{align}
where $\otimes$ is the outer product. It represents the dynamic importance of individual attention traits in the prediction of the current fixation, providing a more profound understanding of individualized visual behavior.

\subsection{Adaptive Fixation Prioritization}
\label{sec:method2}

While the observer-centric feature integration focuses on the fusion of input features, the adaptive fixation prioritization module addresses the variations of gaze behavior at the output end of the decoder. To achieve this, instead of directly predicting fixation positions, our approach, aimed at individualizing fixation predictions, takes a distinct path. We start by extracting semantic feature maps, denoted as $\Vec{A}_t$, from the decoder. These feature maps are subsequently prioritized using attention weights specific to each observer, providing a pragmatic means of refining fixation outputs based on their unique attention traits.

To elaborate on the process, we begin by element-wise multiplication of the semantic feature maps $\Mat{A}_t$ with the input visual features $\Mat{E}$, and then perform average-pooling along the spatial dimensions, resulting in a feature vector that characterizes the observer's attention distribution across different semantic feature channels, defined as 
\begin{equation}
    \Vec{V}_t = \text{AvgPool}_{s}(\Mat{E}\circ \Mat{A}_{t}).
    \label{equ:semantic-vector}
\end{equation}

Considering that the visual preferences of various semantic features may vary for different observers, we introduce normalized attention weights $\Vec{\beta}$ that prioritize the different feature channels, taking into account the observer feature:
\begin{equation}
    \Vec{\beta}_t = {\rm{softmax}}(\Vec{w}_{b}^T \tanh (\Mat{W}_{b} \Vec{V}_t + \Mat{W}_{um} \Vec{u} )),
    \label{equ:unnormalized-attention-weight-semantic-vector}
\end{equation}
where $\Mat{W}_{b}$, $\Mat{W}_{um}$ and $\Vec{w}_{b}$ are learnable parameters. Finally, the attention weights $\Vec{\beta}_t$ are applied to the corresponding semantic feature maps $\Vec{A}_t$ to compute the output
\begin{equation}
    \Vec{m}_t = \Vec{\beta}_t^T\Vec{A}_t.
    \label{equ:final-action-map}
\end{equation}

This mechanism reshapes the scanpath prediction process into a weighted combination of multi-channel feature maps, allowing for the adaptive integration of these maps into the output fixation map. This approach allows the models to refine the fixation positions, providing a precise prediction of an individual's unique scanpath.

\section{Experiments}
\label{sec:experiment}

This section reports comprehensive experimental results and analyses, demonstrating the effectiveness and generalizability of our method across various datasets, model architectures, and visual tasks. For further results, analyses, and implementation details, please refer to the Supplementary Material.

\begin{table*}[t]
\centering
{
\footnotesize
\begin{tabular}{lccc|ccc|ccc|ccc}
  & \multicolumn{3}{c}{OSIE~\cite{juan:2014:osie}}& \multicolumn{3}{c}{OSIE-ASD~\cite{shuo:2015:austim}} & \multicolumn{3}{c}{COCO-Search18~\cite{zhibo:2020:cocosearch}}& \multicolumn{3}{c}{AiR-D~\cite{shi:2020:air}}\\
 \toprule
 Method & SM $\uparrow$ & MM $\uparrow$ & SED $\downarrow$ & SM $\uparrow$ & MM $\uparrow$ & SED $\downarrow$ & SM $\uparrow$ & MM $\uparrow$ & SED $\downarrow$ & SM $\uparrow$ & MM $\uparrow$ & SED $\downarrow$\\ 
\midrule
Human & 0.386 & 0.808 & 7.486 & 0.370 & 0.783 & 7.720 & 0.458 & 0.809 & 1.777 & 0.405 & 0.801 & 7.966\\
\midrule
SaltiNet~\cite{marc:2017:saltinet} & 0.151 & 0.739 & 8.790 & 0.137 & 0.735 & 8.688 & 0.127 & 0.712 & 3.821 &  0.116 & 0.747 & 10.661\\
PathGAN~\cite{marc:2018:pathgan} & 0.056 & 0.744 & 9.393  & 0.042 & 0.732 & 9.342 & 0.231 & 0.714 & 2.454 & 0.072 & 0.739 & 9.888\\
IOR-ROI~\cite{wanjie:2019:iorroi} & 0.294 & 0.791 & 7.966  & 0.301 & 0.788 & 7.655 & 0.197 & 0.787 & 7.087 & 0.239 & 0.791 & 8.584\\
ChenLSTM~\cite{xianyu:2021:vqa} & 0.373 & 0.804 & 7.309 &  0.341 & 0.791 & 7.602 & 0.454 & 0.799 & 1.932 & 0.356 & 0.808 & 7.845\\
Gazeformer~\cite{sounak:2023:gazeformer}  &  0.372 & 0.809 & 7.298 & 0.388 & 0.792 & 7.081 & 0.432 & 0.796 & 2.023 & 0.349 & 0.810 & 8.004 \\
\midrule 
ChenLSTM-FT & 0.378 & 0.808 & 7.344  & 0.394 & 0.796 & 7.067 & 0.454 & 0.804 & 1.936 & 0.341 & 0.806 & 8.282\\
Gazeformer-FT  &0.373 & 0.810 & 7.319  & 0.387 & 0.795 & 7.083 & 0.432 & 0.796 & 2.026 & 0.350 & 0.812 & 8.068\\
ChenLSTM-ISP & 0.377 & 0.810 & 7.284 & 0.401 & \textbf{0.798} & \textbf{6.599} & \textbf{0.480} & \textbf{0.811} & \textbf{1.862} & \textbf{0.371} & 0.813 & \textbf{7.651}\\
Gazeformer-ISP & \textbf{0.390} & \textbf{0.813} & \textbf{7.163} & \textbf{0.406} & 0.797 & 6.823 & 0.455 & 0.806 & 1.997 & 0.362 & \textbf{0.814} & 7.911 \\ 
\bottomrule
\end{tabular}
}

\caption{Comparison of value-based evaluation results for models' ability to predict the scanpaths of individual observers.} 
\label{table:sota-scanpath-rlts}
\end{table*}
\subsection{Experiment Settings}
\label{sec:implementation}

\textbf{Tasks and Datasets.} We conduct experiments on four eye-tracking datasets featuring a variety of visual tasks, including free-viewing, visual search, and visual question answering: \textit{OSIE} \cite{juan:2014:osie} comprising 700 images with free-viewing gaze data from 15 undergraduate and graduate students aged 18–30, \textit{OSIE-ASD} \cite{shuo:2015:austim} with free-viewing gaze data from 20 individuals with ASD and 19 controls, spanning ages 21 to 60, including 33 males and 6 females, \textit{COCO-Search18} \cite{zhibo:2020:cocosearch} (target-present subset) featuring 6202 images with gaze data from 6 males and 4 females aged 18 to 30, collected under a visual search task, and \textit{AiR-D} \cite{shi:2020:air} offering images and questions from the GQA dataset \cite{drew:2019:gqa} with gaze and question-answering data from 16 males and 4 females aged 18 to 38. Dataset splits follow ChenLSTM \cite{xianyu:2021:vqa} for the OSIE, OSIE-ASD, and AiR-D datasets, and the Gazeformer \cite{sounak:2023:gazeformer} for the COCO-Search18.

\textbf{Evaluation Metrics.} 
We conduct individualized scanpath prediction evaluation using two complementary sets of metrics: value-based metrics and ranking-based metrics.
The \textbf{value-based} metrics measure the similarity or dissimilarity between the prediction and ground-truth scanpaths of the same observer. Different from existing studies~\cite{xianyu:2021:vqa} that compare a generic prediction with all observers' ground-truth scanpaths, we evaluate each individualized prediction against the corresponding observer's ground truth. Specifically, \textit{ScanMatch (SM)}~\cite{filipe:2010:scanmatch,hiroyuki:2013:gazeparser} measures the similarity of fixation position and duration using the Needleman-Wunsch algorithm~\cite{saul:1970:needleman-wunsch-algorithm};
\textit{MultiMatch (MM)}~\cite{richard:2012:multimatch} measures scanpath similarity regarding shape, direction, length, position, and duration; \textit{String-Edit Distance (SED)}~\cite{lapo:2020:wave-propagation-attention,stephan:1997:sed,tom:2008:sed} converts scanpaths into strings by associating each image region with a character. To evaluate how well the model predicts distinctly different scanpaths for different observers, we also employ \textbf{ranking-based} metrics. For each predicted scanpath, we rank the ground-truth scanpaths based on their ScanMatch similarity. \textit{Recall at K (R@K)}~\cite{jiacheng:2021:gpo,hao:2019:vse} quantifies whether the correct scanpath (\ie, that from the same observer) is within the top-K most similar scanpaths. \textit{Mean Reciprocal Rank (MRR)}~\cite{abhishek:2017:vd,abhishek:2019:vd,jinhui:2022:visualhow} measures the quality of the ranking by calculating the reciprocal of the rank of the correct scanpath. Thus, the combination of value-based metrics focusing on the specific observer and ranking-based metrics considering all observers offers a comprehensive and robust performance evaluation.

\begin{table*}[t]
\centering
{
\footnotesize
\begin{tabular}{lccc|ccc|ccc|ccc}
 & \multicolumn{3}{c}{OSIE~\cite{juan:2014:osie}}& \multicolumn{3}{c}{OSIE-ASD~\cite{shuo:2015:austim}} & \multicolumn{3}{c}{COCO-Search18~\cite{zhibo:2020:cocosearch}}& \multicolumn{3}{c}{AiR-D~\cite{shi:2020:air}}\\
\toprule
Method & MRR $\uparrow$& R@1 $\uparrow$& R@5 $\uparrow$ & MRR $\uparrow$& R@1 $\uparrow$& R@5 $\uparrow$ & MRR $\uparrow$& R@1 $\uparrow$& R@5 $\uparrow$ & MRR $\uparrow$& R@1 $\uparrow$& R@5 $\uparrow$\\
\midrule
SaltiNet~\cite{marc:2017:saltinet} & 0.213& 5.619 & 32.286 & 0.107 & 2.454 & 12.454  & 0.293 & 10.114 & 49.804 & 0.295 & 10.210 & 49.930\\
PathGAN~\cite{marc:2018:pathgan} & 0.221 & 6.667 & 33.048 & 0.110 & 2.601 & 12.894  & 0.294 & 10.082 & 50.245 & 0.293 & 10.000 & 50.629\\
IOR-ROI~\cite{wanjie:2019:iorroi} & 0.218 & 6.762 & 31.524 & 0.109 & 2.784 & 12.454 & 0.292 & 9.673 & 50.507 &  0.291 & 9.814 & 48.567 \\
ChenLSTM~\cite{xianyu:2021:vqa} & 0.222 & 7.048 & 32.952 & 0.108 & 2.418 & 13.114 & 0.296 & 10.199 & 50.719 & 0.297 & 	9.957 & 51.433\\
Gazeformer~\cite{sounak:2023:gazeformer}  & 0.223 & 7.048 & 32.476 & 0.107 & 2.564 & 11.758 & 0.292 & 9.873 & 50.114 & 0.299 & 10.459 & 51.361\\
\midrule
ChenLSTM-FT &  0.225 & 6.667 & 34.381  & 0.113 & 2.711 & 12.637 &0.298 & 10.641& 49.820 & 0.294 & 10.118 & 50.262 \\
Gazeformer-FT  & 0.217 & 6.000 & 32.857 & 0.108 & 2.528 & 13.223 & 0.293 & 10.183 & 50.000 & 0.300 & 	9.599 & 51.863\\
ChenLSTM-ISP& \textbf{0.291} & \textbf{12.667} & \textbf{44.095} & \textbf{0.147} & \textbf{4.835} & \textbf{19.194} & \textbf{0.369} & \textbf{16.639} & \textbf{61.769} & \textbf{0.338} & \textbf{13.610} & 57.235 \\
Gazeformer-ISP & 0.268 & 10.095 & 41.905 & 0.141 & 4.286 & 18.571  & 0.353  & 15.299  & 60.020 & 0.334  & 13.539 & \textbf{57.450}  \\
\bottomrule
\end{tabular}
}
\caption{Comparison of ranking-based evaluation results for models' ability to distinguish different observers.} 
\label{table:sota-scanpath-retrieval-rlts}
\end{table*}

\begin{table*}[t]
\centering
\resizebox{1\textwidth}{!}{
\begin{tabular}{ccc|cccccc|cccccc}
 \multicolumn{3}{c}{Modules}& \multicolumn{6}{c}{ChenLSTM} & \multicolumn{6}{c}{Gazeformer}\\
\toprule
 OE & FI & FP & SM $\uparrow$& MM $\uparrow$& SED $\downarrow$ & MRR $\uparrow$& R@1 $\uparrow$& R@5 $\uparrow$ & SM $\uparrow$& MM $\uparrow$& SED $\downarrow$ & MRR $\uparrow$& R@1 $\uparrow$& R@5 $\uparrow$\\
\midrule
& & & 0.341 & 0.791 & 7.602 & 0.108 & 2.418 &13.114 & 0.388 & 0.792 & 7.081 & 0.107 & 2.564 & 11.758 \\
\checkmark & & & 0.377 & 0.791 & 7.112 & 0.110 & 2.601 & 13.000 & 0.397 & 0.796 & 7.079 & 0.122 & 3.017 & 15.092\\
\checkmark &  \checkmark & & 0.389 & 0.795 & 7.064 & 0.122 & 3.150 & 15.238  & 0.398 & 0.796 & 6.982 & 0.134 & 3.810 & 17.509 \\
\checkmark &   & \checkmark & 0.389 & 0.795 & 7.063 & 0.112 & 2.784 & 13.150 & 0.397 & 0.797 & 7.073 & 0.120 & 3.077 & 15.165  \\
\checkmark &  \checkmark & \checkmark & \textbf{0.401} & \textbf{0.798} & \textbf{6.599} & \textbf{0.147} & \textbf{4.835} & \textbf{19.194}  & \textbf{0.406} & \textbf{0.797} & \textbf{6.823} & \textbf{0.141} & \textbf{4.286} & \textbf{18.571}\\
\bottomrule 
\end{tabular}
}
\caption{Ablation study for the proposed technical components: observer encoder (OE), observer-centric feature integration (FI), and adaptive fixation prioritization (FP).} 
\label{table:ablation-scanpath-rlts}
\end{table*}

\textbf{Compared Models.} 
We implement two individualized scanpath prediction models representing typical autoregressive and non-autoregressive sequential processing paradigms, respectively: \textit{ChenLSTM-ISP} adapts the ChenLSTM~\cite{xianyu:2021:vqa} model, incorporating the observer encoder and the observer-centric feature integration for input processing. The model's LSTM decoder outputs are further modified for the proposed adaptive fixation prioritization. Similarly, we implement the \textit{Gazeformer-ISP} model upon the Gazeformer~\cite{sounak:2023:gazeformer} architecture. It replaces the original visual-semantic joint embedding with our observer-centric feature integration and changes the Transformer decoder outputs from fixation coordinates to feature maps. We compare these ISP models with their general counterparts and other general scanpath prediction models, including SaltiNet~\cite{marc:2017:saltinet}, PathGAN~\cite{marc:2018:pathgan}, and IOR-ROI~\cite{wanjie:2019:iorroi}. In addition, we fine-tune the general models on individual observer data (\ie, ChenLSTM-FT, Gazeformer-FT) to provide a baseline for assessing the impact of explicitly incorporating observer-specific characteristics.

\textbf{Implementation Details.} We implement ChenLSTM~\cite{xianyu:2021:vqa} and Gazeformer~\cite{sounak:2023:gazeformer} following the original methods, such as using the same visual encoder (\ie,  ResNet-50~\cite{kaiming:2016:resnet}) and task encoder (\ie, RoBERTa~\cite{yinhan:2019:roberta} or AiR-M~\cite{shi:2020:air} or CenterNet~\cite{zhou:2019:objects} object detector). For both models, the number of output feature channels for $\Vec{A}_t$ is empirically set to 4. Specifically, for ChenLSTM~\cite{xianyu:2021:vqa} and Gazeformer~\cite{sounak:2023:gazeformer}, we adopt supervised learning for 15 epochs and self-critical sequence training (SCST)~\cite{xianyu:2021:vqa,steven:2017:scst} for the remaining 10 epochs. In supervised learning, we train our model using the Adam~\cite{diederik:2015:adam} optimizer with learning rate $10^{-4}$ and weight decay $5\times10^{-5}$, while in the SCST, we linearly
decayed learning rates starting at $10^{-5}$. To improve the learning of discriminative features across observers, each training batch includes different scanpaths for the same image.

\subsection{Quantitative Results}
\label{sec:value-based-result}
We present value- and ranking-based evaluation results to assess the effectiveness of our ISP models in capturing the unique attention traits of individual observers.

Table~\ref{table:sota-scanpath-rlts} presents the \textbf{value-based} evaluation results revealing how model predictions resemble the ground truth scanpath of each observer.
While fine-tuning leads to minor improvements in some cases (\eg, OSIE and OSIE-ASD), it struggles on datasets with less distinct inter-observer differences (\eg, COCO-Search18 and AiR-D). In contrast, the ISP models consistently outperform the general methods and fine-tuning, indicating their ability to adapt to the unique attention traits of observers. This is particularly evident in the improved performance (\eg, Gazeformer-ISP, SM=0.406) on the OSIE-ASD dataset with a diverse range of observer demographics. These results suggest that our method, by directly targeting the modeling of observer-specific attention patterns, offers more robust and effective individualization.

Table~\ref{table:sota-scanpath-retrieval-rlts} presents \textbf{ranking-based} evaluation comparing models' ability to distinguish ground-truth scanpaths. 
General models, which are observer-agnostic, cannot differentiate the ground-truth scanpaths from similar ones (\eg, ChenLSTM, R@1=2.4\% on OSIE-ASD, lower than random). Even after fine-tuning with individual eye-tracking data, their performance improvements are marginal (\eg, ChenLSTM-FT, R@1=2.7\% on OSIE-ASD), because independently tuning parameters cannot effectively learn features that distinguish each observer from the others. Differently, the individualized models achieve promising results across all metrics and datasets. From ChenLSTM to ChenLSTM-ISP, R@1 is significantly improved to 4.8\% on the OSIE-ASD dataset, doubling the probability of finding the correct scanpath. It suggests that the ISP models can predict scanpaths that align closely with an observer's unique attention traits. Between network architectures, ChenLSTM-ISP consistently outperforms Gazeformer-ISP when ranking scanpaths. This performance gain may be attributed to LSTM's autoregressive nature which is more effective than Transformer's parallel approach in learning fine-grained spatiotemporal differences.

\subsection{Ablation Study}
To evaluate the significance of the three technical components: observer encoder (OE), observer-centric feature integration (FI), and adaptive fixation prioritization (FP), we conduct an ablation study on the OSIE-ASD dataset~\cite{juan:2014:osie} by applying them incrementally to the ChenLSTM and Gazeformer models. Table~\ref{table:ablation-scanpath-rlts} shows that a fundamental module OE results in a significant improvement in the value-based evaluation and highlights its role of encoding attention traits of observers. Furthermore, based on OE, both FI and FP have notable impacts on the model performance. First, both components achieve similar performance improvements in SM, MM, and SED, demonstrating their ability to improve the overall accuracy of scanpath predictions. Further, regarding the MRR, R@1, and R@5 metrics, FI results in more significant improvements than FP, suggesting that the seamless integration of various input features is more substantial than FP's ability to prioritize where to look at the output end. We also notice that combining both modules leads to the most significant overall performance improvements, indicating that FI and FP offer complementary enhancements. Ablation studies on the other datasets are reported in the Supplementary Material.

\begin{figure}[t]
\centering
\includegraphics[width=1.0\linewidth]{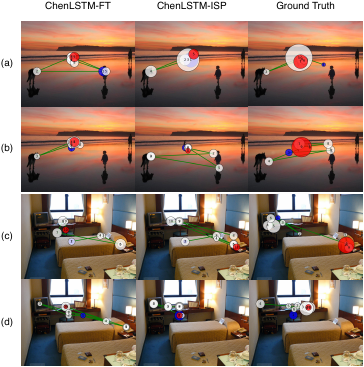}
\caption{Qualitative examples of scanpaths predicted by ChenLSTM-FT, ChenLSTM-ISP, and ground truth. Each row compares the model predictions and the ground truth scanpath of one observer. These observers show different gaze patterns, including (a) focusing on the image center, (b) exploring different people and objects, (c) exploring broadly in the scene, and (d) focusing on a particular region. The \textcolor{blue}{blue} and \textcolor{red}{red} dots indicate the beginning and the end of the scanpath, respectively.}
\label{fig:qual}
\end{figure}

\begin{figure}[t]
\centering
\includegraphics[width=1.0\linewidth]{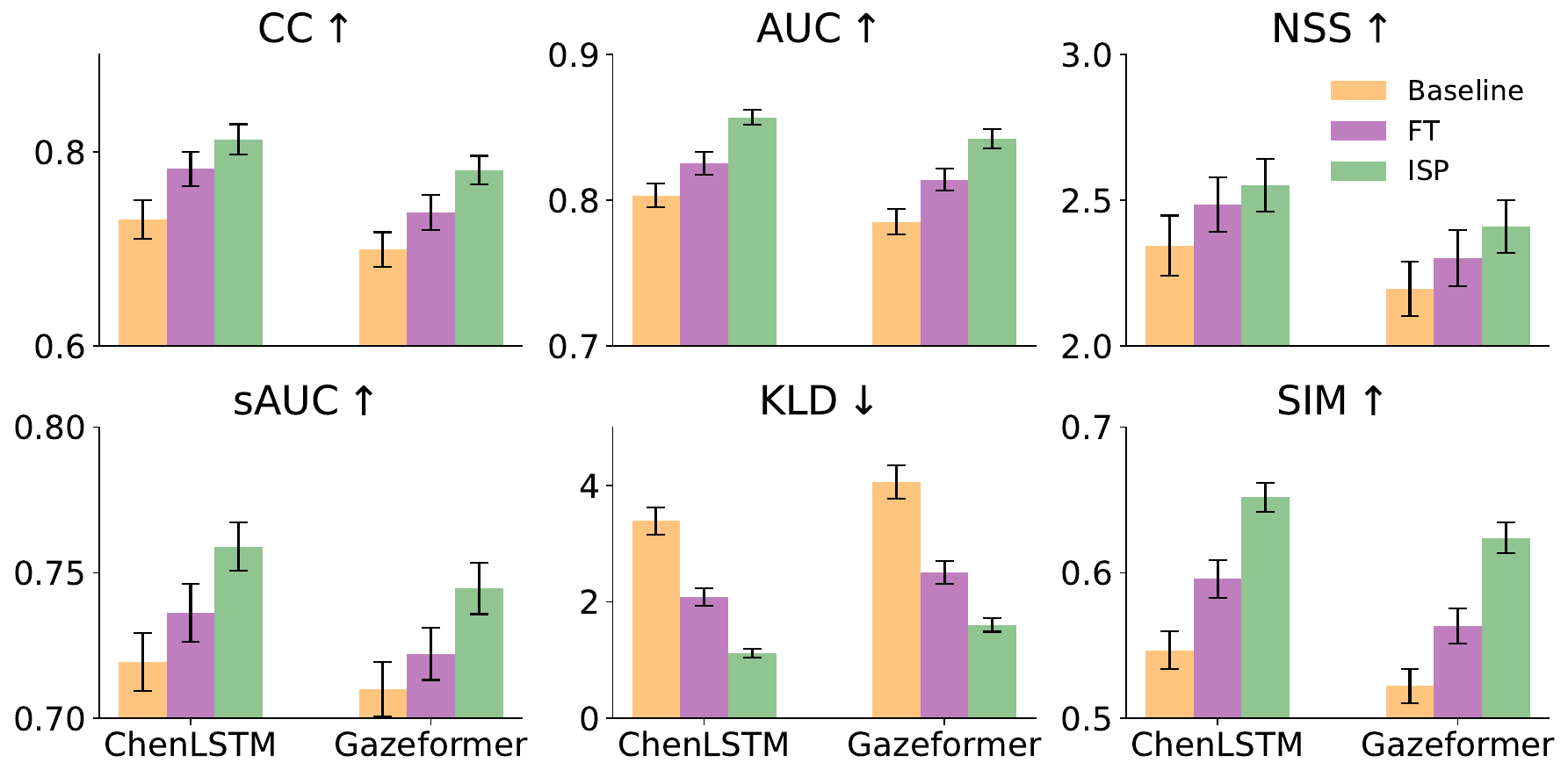}
\caption{Saliency evaluation results of the baselines, fine-tuned (FT) models, and ISP models. Error bars indicate the standard error of the mean.}
\label{fig:saliency_analysis}
\vspace{-1em}
\end{figure}

\subsection{Qualitative Examples}

To understand how the predicted scanpaths align with observer-specific gaze patterns, we present a qualitative comparison in Figure~\ref{fig:qual}. Figure~\ref{fig:qual}a and Figure~\ref{fig:qual}b compare the scanpaths between an observer with autistic traits and a non-autistic observer. It can be seen that observer (a) focused on the center of the image while avoiding direct gaze at people, while observer (b) looked at people more frequently. Figure~\ref{fig:qual}c and Figure~\ref{fig:qual}d compare the scanpaths of two observers responding to the question `What is the device on top of the nightstand made of wood?' with different answers. 
Observer (c) successfully found the correct answer `phone' by searching broadly within the image, but observer (d) responded with an incorrect answer `television' because the fixations were mostly distributed around the television. Notably, while the fine-tuning approach (column 1) falls short in capturing observer-specific gaze patterns, the ISP models' predictions (column 2) better align with the scanpaths of the human observers (column 3). This capability of ISP models opens up new avenues for understanding and interpreting individual differences in visual perception and decision-making processes.

\subsection{From Scanpaths to Saliency Maps}
To further confirm the effectiveness of our ISP method, we assess the spatial accuracy of the predicted fixations using established saliency evaluation metrics~\cite{ming:2015:salicon,xun:2015:salicon}, including Linear Correlation Coefficient (CC), Area Under the ROC curve (AUC), Normalized Scanpath Saliency (NSS), shuffled AUC (sAUC), Kullback-Leibler divergence (KLD), and similarity metric (SIM). Saliency maps are generated by aggregating predicted fixations from all observers and applying a Gaussian kernel smoothing to all fixation points. Figure~\ref{fig:saliency_analysis} shows the substantial improvement of the ISP models over the baselines and fine-tuned models when applied to the OSIE-ASD~\cite{shuo:2015:austim} dataset. This improvement shows that our method not only accurately predicts individual observers' fixations but also enhances the overall prediction of fixation distributions for the population.

\subsection{Semantic Analyses}

Moving forward, we conduct statistical analyses on the OSIE-ASD dataset to test ISP models' ability to learn the attention differences across observers and populations. While the evaluations above focus on fixation positions and durations, this analysis considers how the predicted fixations align with the ground truth regarding their semantic-level statistics. Specifically, we group fixations into three categories based on the region of interest (ROI) annotations provided by  OSIE~\cite{juan:2014:osie}, which are social regions (directly relating to humans, including faces, emotion, touched, gazed), nonsocial regions (\eg, implied motion, relating to nonvisual senses, designed to attract attention, and other objects), and background. Each observer has a unique fixation distribution over the three categories (\ie, social, nonsocial, and background), which enables the following individual-level and population-level analyses.

\begin{table}[t]
\centering
{\small
\begin{tabular}{lccc}
\toprule
Method & Social & Nonsocial & Background\\
\midrule
ChenLSTM~\cite{xianyu:2021:vqa} & 0.181 & -0.159 & 0.067\\
Gazeformer~\cite{sounak:2023:gazeformer}  & -0.141 & -0.253 & -0.211\\
\midrule
ChenLSTM-FT &  0.137 & 0.040 & -0.166 \\
Gazeformer-FT  & 0.045 & 0.164 & 0.051\\
ChenLSTM-ISP& \textbf{0.621} & \textbf{0.655} & \textbf{0.720}\\
Gazeformer-ISP &  \textbf{0.692} & \textbf{0.572} & \textbf{0.699} \\
\bottomrule 
\end{tabular}
}
\caption{Spearmans' correlation coefficients of fixation proportions in 3 semantic ROIs (\ie, social, nonsocial, and background) between the ground truth and predictions. Bold numbers indicate significant positive correlations ($p<0.05$).}
\label{table:individual-analysis}
\end{table}

\begin{figure}[t]
\centering
\includegraphics[width=1.0\linewidth]{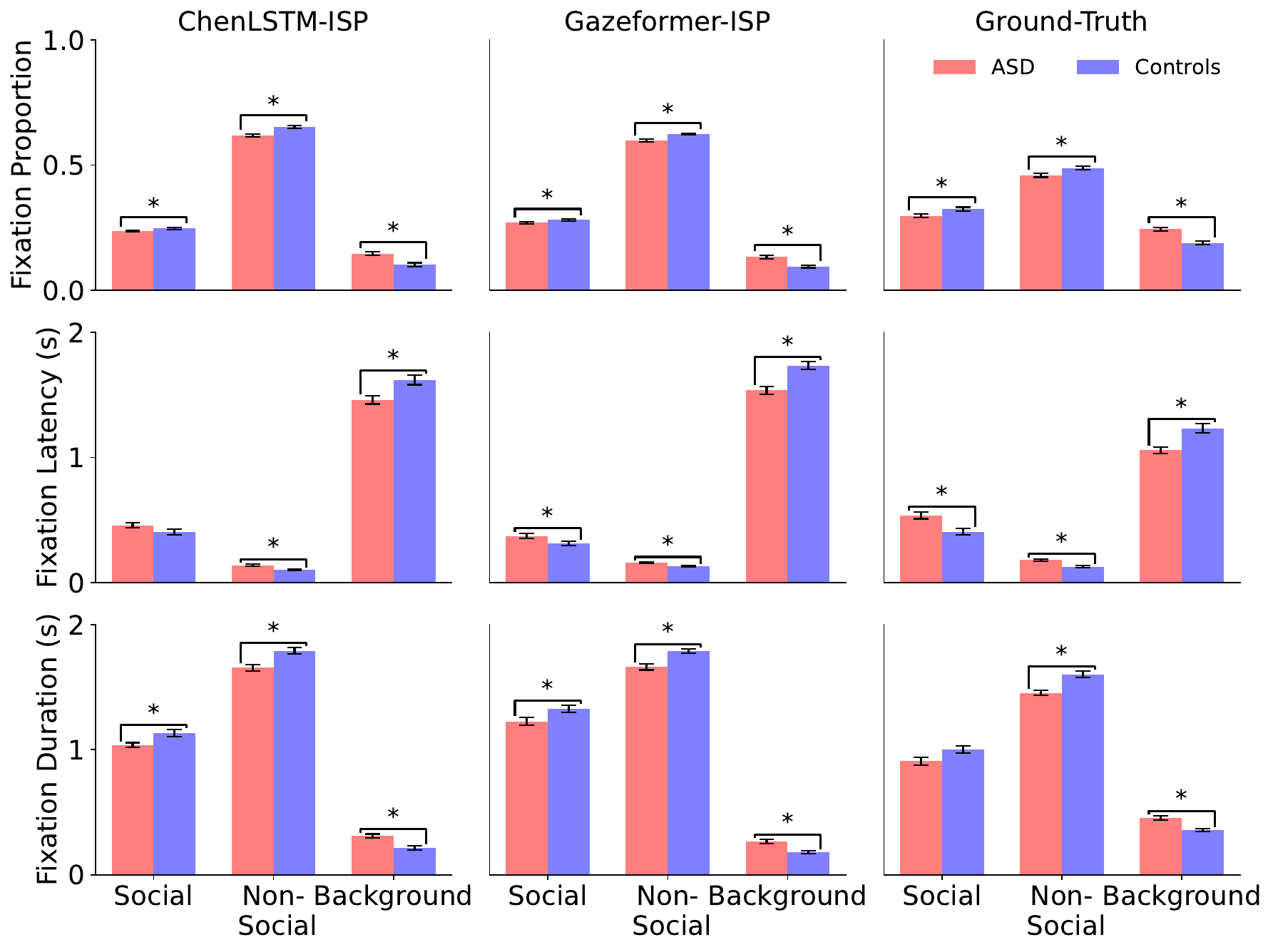}
\caption{Statistical comparison between the predicted fixations for the ASD and Control groups~\cite{shuo:2015:austim}. Error bars indicate the standard error of the mean. Asterisks indicate significant differences (unpaired t-test, $p < 0.05$).}
\label{fig:fixation_analysis}
\end{figure}

\begin{figure}[t]
\centering
\includegraphics[width=1.0\linewidth]{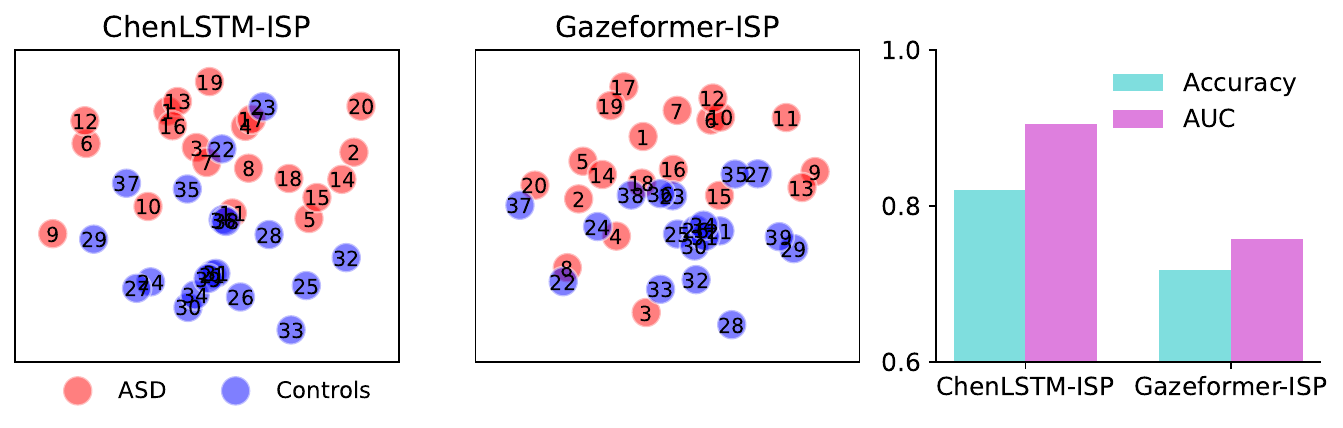}
\caption{Visualization of features extracted from ISP models (numbers indicate observer identities) and results of ASD classification using the features.}
\label{fig:tsne_analysis}
\end{figure}

\textbf{Individual Level.} To evaluate how the predicted scanpaths resemble human fixation statistics, we rank observers by their proportion of fixations in each category. The fixations can be obtained from the model predictions or the ground truth. Table~\ref{table:individual-analysis} presents Spearman's rank correlation coefficient~\cite{alexander:2011:saliency} to compare the observer rankings between the predictions and the ground-truth fixations. While fine-tuning is less effective, showing low correlations across all categories, ISP models consistently achieve significant and high positive correlations, suggesting their ability to resemble each human observer's unique fixation patterns.

\textbf{Population Level.} Beyond individual characterization, ISP models also effectively capture and reproduce distinctive attention traits observed at the population level. For example, individuals with ASD exhibit lower proportions, higher latency, and shorter duration of fixations to both social and nonsocial cues~\cite{shuo:2015:austim,mikle:2005:austim,mark:1998:austim}. Figure~\ref{fig:fixation_analysis} shows that fixations predicted by the ISP models achieve similar statistics. The statistical agreement between the model predictions and the ground-truth scanpaths demonstrates our method's ability to generalize and represent population-level characteristics, reinforcing its potential utility in a variety of applications.

\subsection{Application}

To showcase the potential applicability of ISP models in the diagnosis of neurodevelopmental conditions, we visualize ISP model features and use these features to classify people with ASD. First, the individualization ability of our method is highlighted through t-distributed stochastic neighbor embedding (t-SNE) visualization. By concatenating all observer-specific features from Equations \eqref{equ:attend-map-v-u}, \eqref{equ:ust-vector}, \eqref{equ:uct-vector}, and \eqref{equ:unnormalized-attention-weight-semantic-vector}, into $\Vec{v}=[\Mat{W}_{mu} \Vec{u}\mathbin\Vert\Vec{W}_{us}\Vec{u}\mathbin\Vert\Vec{W}_{uc}\Vec{u}\mathbin\Vert\Mat{W}_{um} \Vec{u}]$, where $\mathbin\Vert$ represents the vector concatenation, Figure~\ref{fig:tsne_analysis} shows that the ISP model features can clearly distinguish people with ASD from the controls. It is noteworthy that such features are learned in an unsupervised manner without knowing each observer's class label, suggesting the strong learning power of the ISP models. Further, based on a leave-one-out cross-validation, we train a two-layer perceptron to classify people with ASD using the extracted feature $\Vec{v}$. ChenLSTM-ISP and Gazeformer-IPS achieve 82.1\% and 71.8\% classification accuracy, respectively, similar to clinical gold standards~\cite{torbjorn:2013:asd,ming:2017:asd}. 
These results demonstrate ISP models' potential in real-world healthcare applications.

\section{Conclusion}
\label{sec:conclusion}
We have introduced a novel approach to predicting individualized human visual scanpaths. Our approach features three novel components: observer encoder, observer-centric feature integration, and adaptive fixation prioritization. Through extensive experiments across multiple datasets, network architectures, and visual tasks, our method consistently outperforms state-of-the-art scanpath prediction methods and individualization based on observer-specific fine-tuning. The results demonstrate the method's ability to generate human-like scanpaths and account for individual observers' gaze patterns. By providing a better understanding of how individuals process visual information, our study has significant implications for tailored, user-centric solutions, such as improving the design of interfaces, products, and services across a wide range of application domains.

\section*{Acknowledgments}

This work is supported by NSF Grant 2143197.

\newpage

{
    \small
    \bibliographystyle{ieeenat_fullname}
    \bibliography{main}
}

\clearpage
\setcounter{page}{1}
\maketitlesupplementary

\section{Introduction}

In the main paper, we have introduced a new method for predicting individual scanpaths (ISP) with three novel components, \ie, observer encoding, observer-centric feature integration, and adaptive fixation prioritization, which can accurately model saccadic eye movements in various tasks, such as free-viewing, visual search, and visual question answering. The experimental results have demonstrated that our method performs competitively and is highly generalizable. This supplementary material supports our main findings with further results and introduces additional implementation details of the proposed method: 
\begin{enumerate}
   \item[1)] Section~\ref{sec:supplementary_method} presents detailed descriptions of how ISP is applied to the Gazeformer architecture and its adaptation for predicting scanpaths in different tasks.
   \item[2)] Section~\ref{sec:supp-result} presents supplementary ablation studies on three other eye-tracking datasets (\ie, OSIE~\cite{juan:2014:osie}, COCO-Search18~\cite{zhibo:2020:cocosearch} and AiR-D~\cite{shi:2020:air}) to demonstrate the effectiveness of the three novel components of our method. Furthermore, we conduct experiments on the OSIE-ASD~\cite{shuo:2015:austim} dataset to investigate the impacts of hyperparameters used in adaptive fixation prioritization. Lastly, we present supplementary experiments to investigate the generalizability of our ISP models on new subjects.
   \item[3)] Section~\ref{sec:supp_saliency} presents additional quantitative results on the saliency map evaluation, comparing our ISP models with different state-of-the-art scanpath prediction approaches and baselines.
   \item[4)] Section~\ref{sec:supp_baseline} presents a quantitative comparison with new baseline models conditioned on the one-hot observer identity, along with statistical analyses to demonstrate our ISP models' individualization ability.
   \item[5)] Section~\ref{sec:supp_qual} presents additional qualitative results, comparing our method with state-of-the-art scanpath prediction approaches. These results highlight the superior performance of our method across three datasets: OSIE-ASD~\cite{juan:2014:osie} (free-viewing), COCO-Search18~\cite{zhibo:2020:cocosearch} (visual search), and AiR-D~\cite{shi:2020:air} (VQA), confirming its adaptability to diverse real-life visual tasks.
\end{enumerate}

\begin{table*}[t]
\centering
{\small
\begin{tabular}{ccc|cccccc|cccccc}
 \multicolumn{3}{c}{Modules}& \multicolumn{6}{c}{ChenLSTM} & \multicolumn{6}{c}{Gazeformer}\\
\toprule
 OE & FI & FP & SM $\uparrow$& MM $\uparrow$& SED $\downarrow$ & MRR $\uparrow$& R@1 $\uparrow$& R@5 $\uparrow$ & SM $\uparrow$& MM $\uparrow$& SED $\downarrow$ & MRR $\uparrow$& R@1 $\uparrow$& R@5 $\uparrow$\\
\midrule
& & &  0.373 & 0.804 & 7.309 & 0.222 & 7.048 & 32.952 & 0.372 & 0.809 & 7.298 & 0.223 & 7.048 & 32.476 \\
\checkmark &  \checkmark & & 0.376 & 0.806 & \textbf{7.271} & 0.282 & 11.333 & 43.143 & 0.389 & 0.810 & 7.164 & 0.264 & 9.619 & 41.238\\
\checkmark &   & \checkmark & 0.377 & 0.807 & 7.299 & 0.229 & 7.238 & 33.143 &   0.384 & 0.810 & 7.186 & 0.241 & 7.905 & 37.524 \\
\checkmark &  \checkmark & \checkmark & \textbf{0.377} & \textbf{0.810} & 7.284 & \textbf{0.291} & \textbf{12.667} & \textbf{44.095} & \textbf{0.390} & \textbf{0.813} & \textbf{7.163 }& \textbf{0.268} & \textbf{10.095} & \textbf{41.905}\\
\bottomrule 
\end{tabular}
}
\caption{Ablation study for the proposed technical components: observer encoder (OE), observer-centric feature integration (FI) and adaptive fixation prioritization (FP) for OSIE~\cite{juan:2014:osie} dataset.}
\label{table:ablation-scanpath-osie-rlts}
\end{table*}

\begin{table*}[t]
\centering
{\small
\begin{tabular}{ccc|cccccc|cccccc}
 \multicolumn{3}{c}{Modules}& \multicolumn{6}{c}{ChenLSTM} & \multicolumn{6}{c}{Gazeformer}\\
\toprule
 OE & FI & FP & SM $\uparrow$& MM $\uparrow$& SED $\downarrow$ & MRR $\uparrow$& R@1 $\uparrow$& R@5 $\uparrow$ & SM $\uparrow$& MM $\uparrow$& SED $\downarrow$ & MRR $\uparrow$& R@1 $\uparrow$& R@5 $\uparrow$\\
\midrule
& & &  0.454 & 0.799 & 1.932 & 0.296 & 10.199 & 50.719 & 0.432 & 0.796 & 2.023 & 0.292 & 9.873 & 50.114\\
\checkmark &  \checkmark & & 0.471 & 0.809 & 1.896 & 0.365 & 15.266 & 61.360 & 0.450 & 0.804 & 2.021 & 0.347 & 14.678 & 59.170\\
\checkmark &   & \checkmark &   0.474 & 0.807 & 1.871 & 0.329 & 12.994 & 55.574 & 0.450 & 0.799 & 2.014 & 0.321 & 12.096 & 55.149 \\
\checkmark &  \checkmark & \checkmark & \textbf{0.480} & \textbf{0.811} & \textbf{1.862} & \textbf{0.369} & \textbf{16.639} & \textbf{61.769} & \textbf{0.455} & \textbf{0.806} & \textbf{1.997} & \textbf{0.353} & \textbf{15.299} & \textbf{60.020}\\
\bottomrule 
\end{tabular}
}
\caption{Ablation study for the proposed technical components: observer encoder (OE), observer-centric feature integration (FI) and adaptive fixation prioritization (FP) for COCO-Search18~\cite{zhibo:2020:cocosearch} dataset.}
\label{table:ablation-scanpath-cocosearch-rlts}
\end{table*}

\begin{table*}[t]
\centering
{\small
\begin{tabular}{ccc|cccccc|cccccc}
 \multicolumn{3}{c}{Modules}& \multicolumn{6}{c}{ChenLSTM} & \multicolumn{6}{c}{Gazeformer}\\
\toprule
 OE & FI & FP & SM $\uparrow$& MM $\uparrow$& SED $\downarrow$ & MRR $\uparrow$& R@1 $\uparrow$& R@5 $\uparrow$ & SM $\uparrow$& MM $\uparrow$& SED $\downarrow$ & MRR $\uparrow$& R@1 $\uparrow$& R@5 $\uparrow$\\
\midrule
& & & 0.356 & 0.808 & 7.845 & 0.297 & 9.957 & 51.433 & 0.349 & 0.810 & 8.004 & 0.299 & 10.459 & 51.361\\
\checkmark &  \checkmark & & 0.357 &  0.808 & 7.824 & 0.330 & 12.607 & 57.020 &  0.353 & 0.814 & 7.992 & 0.316 & 12.536 & 53.224\\
\checkmark &   & \checkmark & 0.370 & 0.812 & 7.768 & 0.306 & 11.175 & 51.146 & 0.355 & 0.811 & 7.956 & 0.299 & 10.888 & 51.576\\
\checkmark &  \checkmark & \checkmark & \textbf{0.371} & \textbf{0.813} & \textbf{7.651} & \textbf{0.338} & \textbf{13.610} & \textbf{57.235} & \textbf{0.362} & \textbf{0.814} & \textbf{7.911} & \textbf{0.334} & \textbf{13.539} & \textbf{57.450}\\
\bottomrule 
\end{tabular}
}
\caption{Ablation study for the proposed technical components: observer encoder (OE), observer-centric feature integration (FI) and adaptive fixation prioritization (FP) for AiR-D~\cite{shi:2020:air} dataset.}
\label{table:ablation-scanpath-air-rlts}
\end{table*}

\section{Implementation Details of Gazeformer-ISP}
\label{sec:supplementary_method}

We have implemented our ISP on two baseline architectures, ChenLSTM and Gazeformer. The ISP method designed for ChenLSTM has been introduced in Section~\ref{sec:methodology} of the main paper. In this section, we elaborate detailed implementation of ISP on the Gazeformer model. There are three distinctions compared with the ChenLSTM architecture:

\textbf{Task Encoder.} Different from ChenLSTM that uses the machine attention of a VQA model~\cite{shi:2020:air} or object detection outputs to initially guide the first fixation to direct the eye movement based on the given visual task, we first extract the target feature or question feature using the language model RoBERTa~\cite{yinhan:2019:roberta} as $\Vec{v}_{\ell}$. To model the interaction between the visual feature $\Vec{E}$ and the language feature $\Vec{v}_{\ell}$, a task guidance map can be computed through a linear combination:
\begin{equation}
    \Vec{m}_0 = {\rm{softmax}}\big(\Vec{w}_{e\ell}^T \tanh (\Mat{W}_{e\ell} \Vec{E} + \Mat{W}_{m\ell} \Vec{v}_{\ell} )\big),
    \label{equ:attend-task-guidance-map}
\end{equation}
where $\Vec{w}_{e\ell}$, $\Mat{W}_{e\ell}$, $\Mat{W}_{m\ell}$ are learnable parameters. This observer guidance localizes salient image regions of specific interest to the observer.

\textbf{Observer-Centric Feature Integration.} 
Different from the LSTM architecture, Transformers process input sequences in parallel, relying on positional encodings to impart positional information. As a result, the temporal subscript $t$ becomes less relevant in the Transformer context. Therefore, we always set $\Vec{m}_{t-1}$ as $\Vec{m}_{0}$ from Equation~\eqref{equ:attend-task-guidance-map} and eliminate all the $t$ in the subscript of each variables, such as $\Vec{X}_t$, $\Vec{X}_{ut}$, $\Vec{u}_{t}^s$, $\Vec{u}_{t}^c$ and $\Vec{R}_t$.
This simplification maintains consistency within the Transformer architecture, facilitating efficient parallelization and computation.

\textbf{Adaptive Fixation Prioritization.} The original Gazeformer~\cite{sounak:2023:gazeformer} model comprises two variants: the Gazeformer-Reg produces fixation coordinates and is trained using a regression objective, while the Gazeformer-noReg generates a fixation map and is trained through grid-based classification. Our ISP is based on the latter implementation by applying the proposed adaptive fixation prioritization model, using a cross-attention mechanism~\cite{jacob:2018:bert,sounak:2023:gazeformer} to compute the semantic feature maps $\Vec{A}_t$.

\section{Supplementary Ablation Studies}\label{sec:supp-result}

In this section, we conduct comprehensive ablation studies on the OSIE~\cite{juan:2014:osie}, COCO-Search18~\cite{zhibo:2020:cocosearch} and AiR-D~\cite{shi:2020:air} datasets, as well as the ablation study of the number of output feature channels for the proposed adaptive fixation prioritization. Lastly, we discuss how the ISP models can generalize on new subjects and how much gaze data of the new subjects is needed to achieve comparable performance.

\subsection{Ablation Study on OSIE, COCO-Search18 and AiR-D Datasets}
\label{sec:ablation-three-dataset}

In Table~\ref{table:ablation-scanpath-rlts} of the main paper, we have presented a comprehensive ablation study on the OSIE-ASD~\cite{shuo:2015:austim} dataset to evaluate the significance of the proposed technical components: object encoder (OE), observer-centric feature integration (FI), and adaptive fixation prioritization (FP). In this section, to demonstrate the generalizability of our proposed ISP, we further provide more insights on the effectiveness of these proposed technical components in the other eye-tracking datasets (\ie, OSIE~\cite{juan:2014:osie}, COCO-Search18~\cite{zhibo:2020:cocosearch} and AiR-D~\cite{shi:2020:air}), see Tables~\ref{table:ablation-scanpath-osie-rlts},~\ref{table:ablation-scanpath-cocosearch-rlts} and~\ref{table:ablation-scanpath-air-rlts}. Similar to Table~\ref{table:ablation-scanpath-rlts} demonstrating that the FI and FP play a complementary role in leading to the most significant overall performance improvements on different neural network architectures, These supplementary results emphasize the consistent importance of these technical components across diverse scenarios and tasks. This ablation study reinforces the robustness and effectiveness of these components in our proposed ISP method.

\begin{table*}[t]
\centering
{\small
\begin{tabular}{ccccccc|cccccc}
& \multicolumn{6}{c}{ChenLSTM} & \multicolumn{6}{c}{Gazeformer}\\
\toprule
 $L$ & SM $\uparrow$& MM $\uparrow$& SED $\downarrow$ & MRR $\uparrow$& R@1 $\uparrow$& R@5 $\uparrow$ & SM $\uparrow$& MM $\uparrow$& SED $\downarrow$ & MRR $\uparrow$& R@1 $\uparrow$& R@5 $\uparrow$\\
\midrule
1 & 0.389 & 0.795 & 7.064 & 0.122 & 3.150 & 15.238 & 0.398 & 0.796 & 6.982 & 0.134  & 3.810 & 17.509 \\
2 & 0.393 & 0.796 & 6.885 & 0.147 & 4.835 & \textbf{19.414} & 0.406 & \textbf{0.798} & 6.992 & 0.138 & 3.626 & 18.608 \\
4 & \textbf{0.401} & \textbf{0.798} & \textbf{6.599} & \textbf{0.147} & \textbf{4.835} & 19.194 & \textbf{0.406} & 0.797 & \textbf{6.823} & \textbf{0.141}  & \textbf{4.286} & \textbf{18.571}\\
6 & 0.385 & 0.796 & 7.043 & 0.140 & 3.883 & 18.352 & 0.400 & 0.793 & 6.849 & 0.135 & 3.626 & 18.242\\
\bottomrule 
\end{tabular}
}
\caption{Ablation study of different values of hyperparameter $L$ on the OSIE-ASD~\cite{shuo:2015:austim} dataset. We select the best hyperparameter based on the ScanMatch scores.}
\label{table:ablation-map-number-scanpath-rlts}
\end{table*}

\subsection{Ablation Study of Output Feature Maps}

As outlined in the main paper, our adaptive fixation prioritization dynamically integrates a number of output feature maps for predicting the next fixation position. Thus, the number of feature maps, denoted as $L$, is cruicial to the prediction performance. Renowned for its extensive participant pool and diverse demographics, OSIE-ASD~\cite{shuo:2015:austim} provides an ideal testbed for assessing individualized visual scanpath prediction due to its capacity to differentiate among various individuals.
Table~\ref{table:ablation-map-number-scanpath-rlts} reports model performances across diverse configurations on the OSIE-ASD~\cite{shuo:2015:austim} dataset. On the one hand, a lower $L$ (\eg, $L=1$ or $L=2$) leads to feature maps with limited semantic variations, making it challenging to discern distinct individual fixation patterns, resulting in suboptimal outcomes for individual scanpath prediction. On the other hand, a higher $L$ introduces higher dimensionality, leading to overfitting on the training fixation data (\eg, $L=6$), causing a notable decline in evaluation metric scores. Setting $L=4$ results in a reasonable balance, facilitating the learning of diverse fixation semantic feature maps.

\subsection{Cross-Subject Generalizability}
\begin{table*}[t]
\centering
{\small
\begin{tabular}{l|cccccc|cccccc}
\multicolumn{1}{c}{} & \multicolumn{6}{c}{ChenLSTM} & \multicolumn{6}{c}{Gazeformer}\\
\toprule
  Num. of Img. & SM $\uparrow$& MM $\uparrow$& SED $\downarrow$ & MRR $\uparrow$& R@1 $\uparrow$& R@5 $\uparrow$ & SM $\uparrow$& MM $\uparrow$& SED $\downarrow$ & MRR $\uparrow$& R@1 $\uparrow$& R@5 $\uparrow$\\
\midrule
Full OSIE & \textbf{0.377} & \textbf{0.810} & 7.284 & \textbf{0.291} & \textbf{12.667} & 44.095 & \textbf{0.390} & \textbf{0.813} & \textbf{7.163} & 0.268 & 10.095 & 41.905\\
\midrule
20 images & 0.344 & 0.800 & 7.571 & 0.240 & 7.333 & 39.238 & 0.369 & 0.809 & 7.583 & 0.253 & 8.857 & 40.191\\
50 images & 0.358 & 0.805 & 7.409 & 0.259 & 9.143 & 26.191 & 0.365 & 0.809 & 7.370 & 0.264 & 10.571 & 40.667\\
100 images & 0.361 & 0.805 & 7.257 & 0.266 & 10.095 & 40.667 & 0.372 & 0.810 & 7.268 & 0.274 & 11.333 & 40.667\\
200 images & 0.365 & 0.807 & 7.260 & 0.270 & 10.095 & 42.286 & 0.374 & \textbf{0.813} & 7.233 & 0.271 & 10.286 & 43.429\\
560 images & 0.373 & 0.806 & \textbf{7.218} & 0.279 & 10.762 & \textbf{44.476} & 0.376 & 0.811 & 7.325 & \textbf{0.290} & \textbf{12.571} & \textbf{45.714}\\
\bottomrule 
\end{tabular}
}
\caption{Ablation study of the number of gaze data from new subjects.}
\label{table:sup-generalization-rlts}
\end{table*}

To explore the generalizability of our ISP models on new subjects, we conduct fine-tuning experiments. ISP models pre-trained on the OSIE-ASD~\cite{shuo:2015:austim} dataset are fine-tuned on the OSIE~\cite{juan:2014:osie} dataset using varying amounts of gaze data ($N=20, ~50, ~100,~200,~560$ samples per subject). These fine-tuned models are compared to models trained entirely on OSIE data (\ie, Full OSIE). As shown in Table~\ref{table:sup-generalization-rlts}, even with a small number of training samples of new subjects (\eg,  $N=100$), the evaluation results closely resemble those of the Full OSIE model (\eg, Gazeformer-ISP, SM=0.372 on 100 images compared to SM=0.390 from Full OSIE models), demonstrating that models pre-trained on different subjects provide common knowledge for generalizing to new ones. On the other hand, with an increase in the number of images, performance gradually improves and approaches that of the Full OSIE model, indicating that increasing the data can enhance the model's capabilities.

\section{Saliency Evaluations}
\label{sec:supp_saliency}
In this section, we investigate the spatial accuracy of predicted fixations with a comprehensive saliency comparison across a variety of state-of-the-art models and baselines on the OSIE-ASD~\cite{shuo:2015:austim} dataset.

Supplementing the individualized scanpath evaluation results in Table~\ref{table:sota-scanpath-rlts}, we evaluate the spatial accuracy of predicted fixations through population-level saliency maps. In particular, we aggregate the fixations from the generated scanpaths of all observers, and post-process the fixations to obtain smoothed saliency maps~\cite{xiangjie:2023:scandmm}. This analysis compares ISP to state-of-the-art methods and baselines on their ability to capture the overall distribution of fixations across observers. Besides the models presented in Table~\ref{table:sota-scanpath-rlts}, we also add the Le Meur \etal~\cite{olivier:2015:saccadicmodel}, Li \etal~\cite{anqi:2017:individualscanpath}, and DeepGaze III~\cite{matthias:2022:deepgaze} models into the comparison. Table~\ref{table:sup-saliency-rlts} presents the evaluation results. Notably, ISP models outperform both baselines and the previous state-of-the-art method (with the highest average ranking $\bar{R}$) on the OSIE-ASD~\cite{shuo:2015:austim} dataset. This notable improvement confirms the effectiveness of our proposed ISP method in capturing the overall prediction of fixation distributions for the population.
\begin{table*}[t]
\centering
{\small
\begin{tabular}{l|cccccc|c}
\toprule
 Method & CC $\uparrow$ & AUC $\uparrow$& NSS $\uparrow$ & sAUC $\uparrow$& KLD $\downarrow$& SIM $\uparrow$ & $\bar{R}$ $\downarrow$\\
\midrule
SaltiNet~\cite{marc:2017:saltinet} & 0.224 & 0.661 & 0.541 & 0.587 & 1.496 & 0.414 & 10.33\\
PathGAN~\cite{marc:2018:pathgan} & 0.239 & 0.599 & 0.560 & 0.505 & 10.703 & 0.225 & 11.67\\
IOR-ROI~\cite{wanjie:2019:iorroi} & 0.572 & 0.803 & 1.562 & 0.669 & 1.841 & 0.534 & 9.00\\
Le Meur~\etal~\cite{olivier:2015:saccadicmodel} & 0.621 & 0.819 & 1.768 & \textbf{0.752} & 0.702 & 0.542 & 5.42\\
Li~\etal~\cite{anqi:2017:individualscanpath} & 0.538 & 0.854 & 1.767 & 0.686 & 1.409 & 0.542 & 6.92\\
DeepGaze III~\cite{matthias:2022:deepgaze} & 0.786 & \textbf{0.877} & 2.050 & 0.749 & \textbf{0.359} & \textbf{0.694} & 2.33\\
ChenLSTM~\cite{xianyu:2021:vqa} & 0.728 & 0.810 & 2.348 & 0.720 & 3.310 & 0.549 & 6.83\\
Gazeformer~\cite{sounak:2023:gazeformer} & 0.689 & 0.790 & 2.179 & 0.708 & 4.135 & 0.520 & 8.67\\
\midrule
ChenLSTM-FT & 0.782 & 0.830 & 2.495 & 0.736 & 2.151 & 0.599 & 4.33\\
Gazeformer-FT & 0.733 & 0.816 & 2.303 & 0.721 & 2.762 & 0.565 & 6.17\\
ChenLSTM-ISP & \textbf{0.807} & 0.855 & \textbf{2.529} & 0.748 & 1.442 & 0.642 & \textbf{2.17}\\
Gazeformer-ISP & 0.779 & 0.842 & 2.396 & 0.733 & 1.771 & 0.620 & 4.17\\
\bottomrule 
\end{tabular}
}
\caption{Comparison of saliency evaluation results of state-of-the-art models and baselines. $\bar{R}$ indicates the average ranking across all the saliency evaluations.}
\label{table:sup-saliency-rlts}
\end{table*}

\section{Supplementary Baseline Models}
\label{sec:supp_baseline}
In this section, we introduce supplementary baseline models in addition to the fine-tuned (FT) baseline in the main paper. We present a comprehensive comparison of value-based and ranking-based evaluations with the state-of-the-art methods and the baselines on the OSIE-ASD~\cite{shuo:2015:austim} dataset. We also conduct a comprehensive statistical analysis of the attention traits at the population level among the baselines.

\begin{table*}[t]
\centering
{\small
\begin{tabular}{l|cccccc}
\toprule
 Method & SM $\uparrow$& MM $\uparrow$& SED $\downarrow$ & MRR $\uparrow$& R@1 $\uparrow$& R@5 $\uparrow$\\
\midrule
Human & 0.370 & 0.783 & 7.720 &  - & -& - \\
\midrule
SaltiNet~\cite{marc:2017:saltinet} &  0.137 &  0.735 &  8.688 &  0.107 &  2.454 &  12.454\\
PathGAN~\cite{marc:2018:pathgan} &  0.042  & 0.732  & 9.342  & 0.110  & 2.601 &  12.894\\
IOR-ROI~\cite{wanjie:2019:iorroi} & 0.301  & 0.788 &  7.655 &  0.109 &  2.784 &  12.454\\
ChenLSTM~\cite{xianyu:2021:vqa} & 0.341 &  0.791  & 7.602  & 0.108  & 2.418  & 13.114\\
Gazeformer~\cite{sounak:2023:gazeformer} & 0.388 &  0.792  & 7.081 &  0.107 &  2.564  & 11.758\\
\midrule
ChenLSTM-FT & 0.394  & 0.796 &  7.067 &  0.113 &  2.711  & 12.637\\
Gazeformer-FT & 0.387  & 0.795 &  7.083 &  0.108 &  2.528  & 13.223 \\
ChenLSTM-onehot &  0.366 & 0.785 & 7.291 & 0.106 & 2.491 & 11.722\\
Gazeformer-onehot & 0.395 & 0.797 & 7.006 & 0.116 & 2.894 & 14.029 \\
ChenLSTM-ISP & 0.401  & \textbf{0.798}  & \textbf{6.599}  & \textbf{0.147}  & \textbf{4.835} &  \textbf{19.194}\\
Gazeformer-ISP & \textbf{0.406}  & 0.797  & 6.823  & 0.141 &  4.286  & 18.571 \\
\bottomrule 
\end{tabular}
}
\caption{Comparison of value-based evaluation results for models' ability to predict the scanpaths of individual observers and ranking-based evaluation results for models' ability to distinguish different observers.}
\label{table:sup-sota-rlts}
\end{table*}

\subsection{Implementation Details}
While fine-tuning the general models on individual observer data (\ie, ChenLSTM-FT, Gazeformer-FT) provides a baseline for
assessing the impact of explicitly incorporating observer-specific characteristics, we additionally incorporate supplementary baseline models (\ie, ChenLSTM-onehot, Gazeformer-onehot) with the scanpath prediction conditioned on the one-hot observer identity (\eg, concatenating one-hot encoding of the identity with the feature maps to the scanpath decoder). On the one hand, for the fine-tuned models, we fine-tune the pre-trained model on the population gaze data with a reduced learning rate $10^{-5}$ in 2 epochs. On the other hand, for the one-hot baseline models, we train the model similarly to the ISP models mentioned in the main paper, with 15 epochs of supervised learning and 10 epochs of self-critical sequence training (SCST)~\cite{xianyu:2021:vqa,steven:2017:scst} with learning rate $10^{-4}$.

\subsection{Quantitative Results}
In Table~\ref{table:sota-scanpath-rlts} and Table~\ref{table:sota-scanpath-retrieval-rlts} of the main paper, we have conducted a comprehensive comparison of value-based and ranking-based evaluations on the OSIE~\cite{juan:2014:osie}, OSIE-ASD~\cite{shuo:2015:austim} COCO-Search~\cite{zhibo:2020:cocosearch} and AiR-D~\cite{shi:2020:air} datasets to demonstrate the effectiveness of the proposed ISP. Here, we further include the performance on the one-hot baseline models on OSIE-ASD~\cite{shuo:2015:austim} dataset in Table~\ref{table:sup-sota-rlts}. 
Interestingly, the one-hot baseline outperforms the fine-tuned Gazeformer model (Gazeformer-FT) but underperforms the fine-tuned ChenLSTM model (ChenLSTM-FT), yet both baseline methods underperform the proposed ISP approach. Overall, these results highlight ISP's ability to achieve robust and effective individualization by explicitly capturing observer-specific attention patterns.

\subsection{Population-Level Comparison}
\begin{figure*}[t]
\centering
\includegraphics[width=1\linewidth]
{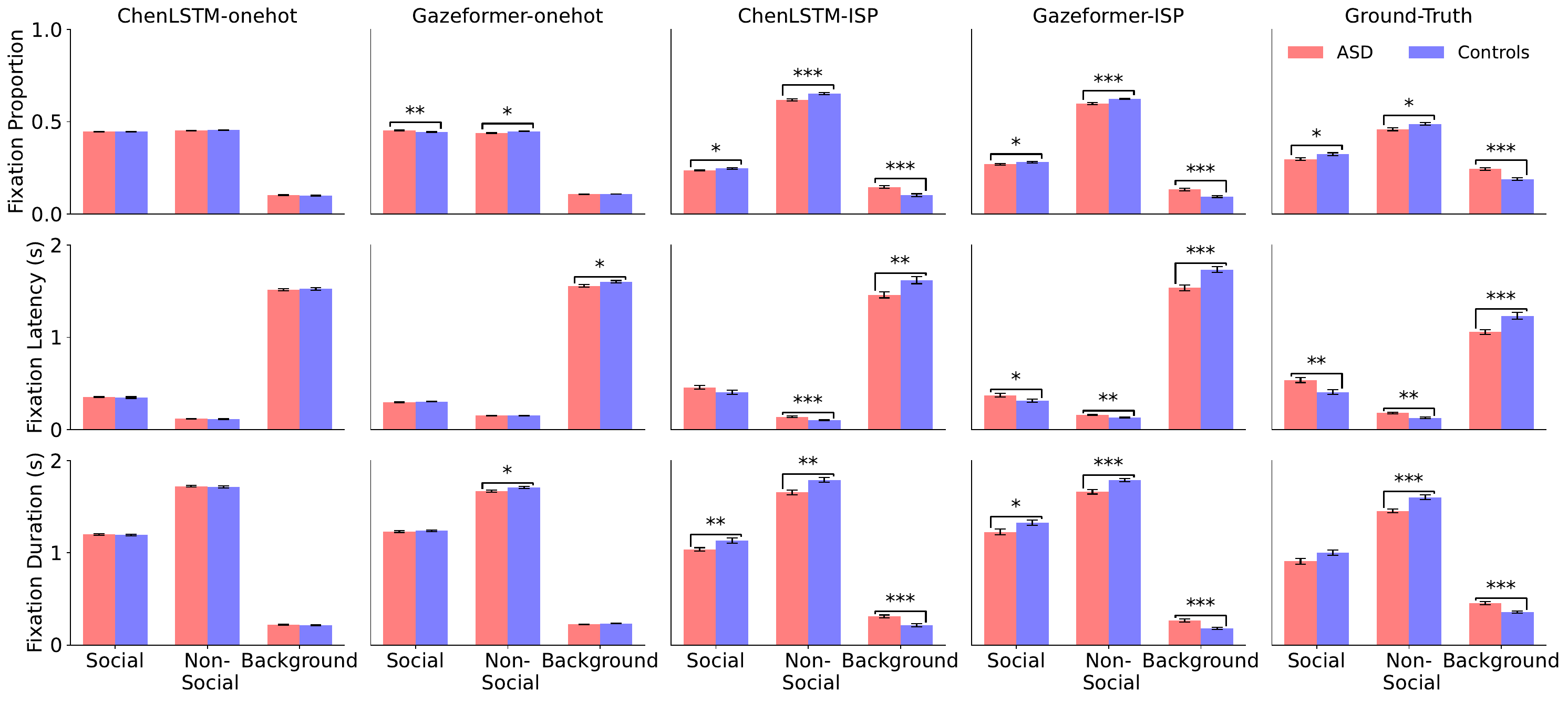}
\caption{Statistical comparison between the predicted fixations for the ASD and Control groups~\cite{shuo:2015:austim} with the baseline. Error bars indicate the standard error of the mean. 
Asterisks indicate significant differences (unpaired t-test,  *: $p<0.05$, **: $p<0.01$, and ***: $p<0.001$).}
\label{fig:fixation_analysis_population_models}
\end{figure*}

We extend the statistical comparison between the predicted fixations
for the ASD and Control groups by incorporating one-hot baseline models (ChenLSTM-onehot, Gazeformer-onehot) into the analysis alongside ISP models and ground truth fixations (Figure~\ref{fig:fixation_analysis_population_models}). 
As shown in Figure~\ref{fig:fixation_analysis_population_models}, the one-hot baselines exhibit significant limitations. ChenLSTM-onehot completely fails to differentiate gaze statistics between subject groups, resulting in no statistically significant differences. Gazeformer-onehot, while partially successful, misses four key statistical differences and even produces one incorrect result, suggesting a higher fixation proportion for individuals with ASD on social semantics, which is a finding contradicted by the data. In contrast, both ChenLSTM-ISP and Gazeformer-ISP models closely resemble the ground truth, accurately capturing the population-level gaze patterns. This success demonstrates the effectiveness of ISP methods in generalizing and representing population-level characteristics.

\section{Supplementary Qualitative Results}
\label{sec:supp_qual}
In addition to the qualitative examples shown in Figure~\ref{fig:qual} of our main paper, we offer an expanded exploration of the qualitative results derived from our ISP method. This extended presentation involves a thorough comparison among ISP models, fine-tuned models, and human ground truth,  covering a spectrum of visual tasks based on the OSIE-ASD~\cite{shuo:2015:austim}, COCO-Search18~\cite{zhibo:2020:cocosearch}, and AiR-D~\cite{shi:2020:air}. Our ISP models consistently align accurately with the scanpaths of individual observers. These qualitative examples demonstrate their potential as a promising and interpretable tool in unraveling visual perception and decision-making processes.

\begin{figure*}[h]
\centering
\includegraphics[width=1.0\linewidth]{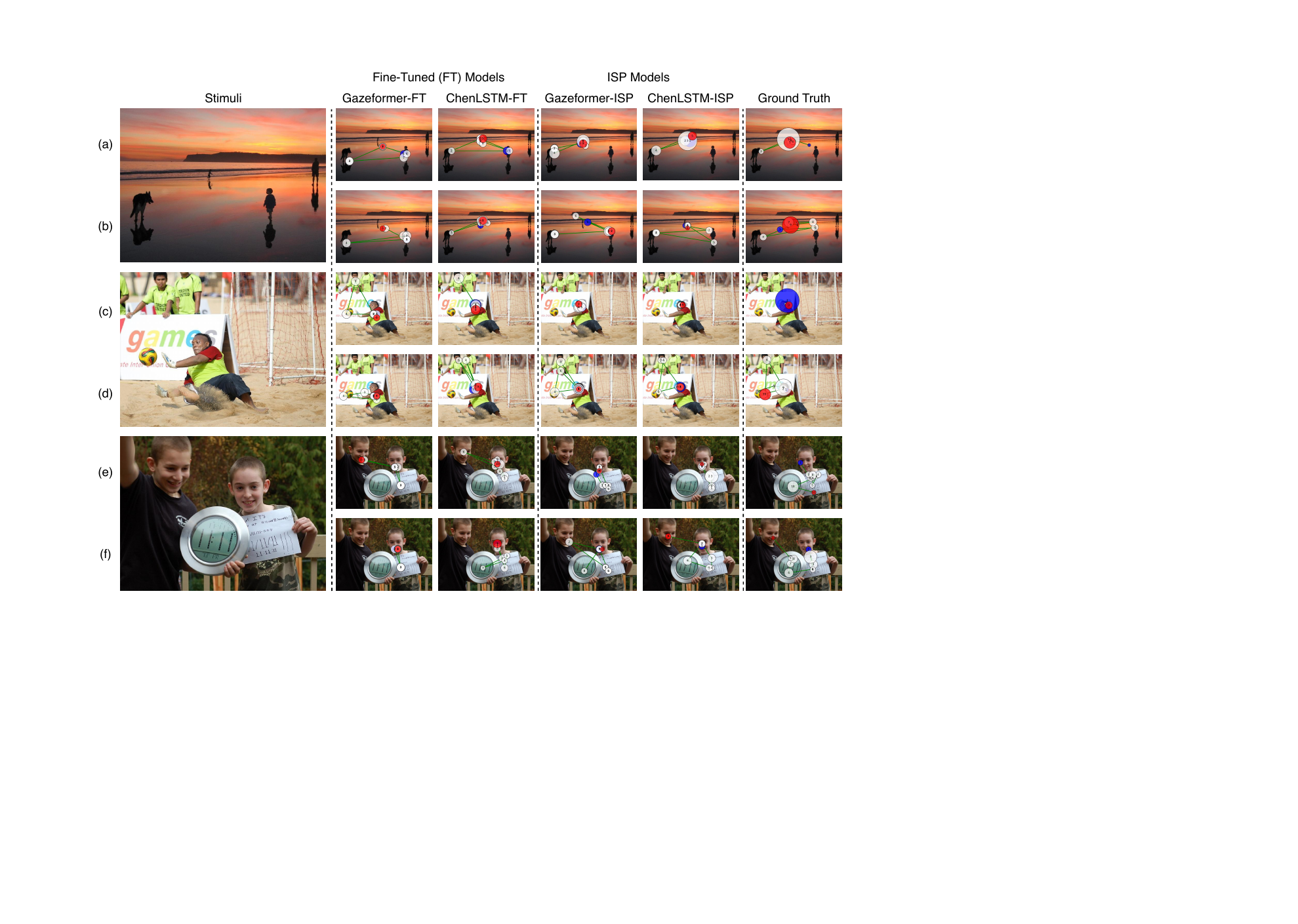}
\caption{Qualitative scanpath examples from Gazeformer-FT, ChenLSTM-FT, Gazeformer-ISP, ChenLSTM-ISP, and ground truth on OSIE-ASD~\cite{shuo:2015:austim}. The first column shows the stimuli, and each row compares model predictions with one observer's ground truth. Observers with ASD exhibit atypical gaze patterns: (a, c) prolonged fixation on central objects, (e) local exploration, while the controls (b, d, f) explore diverse people and objects. \textcolor{blue}{Blue} and \textcolor{red}{red} dots mark the start and end of scanpaths, respectively.}
\label{fig:sup_autism_qual}
\end{figure*}

\begin{figure*}[h]
\centering
\includegraphics[width=1.0\linewidth]{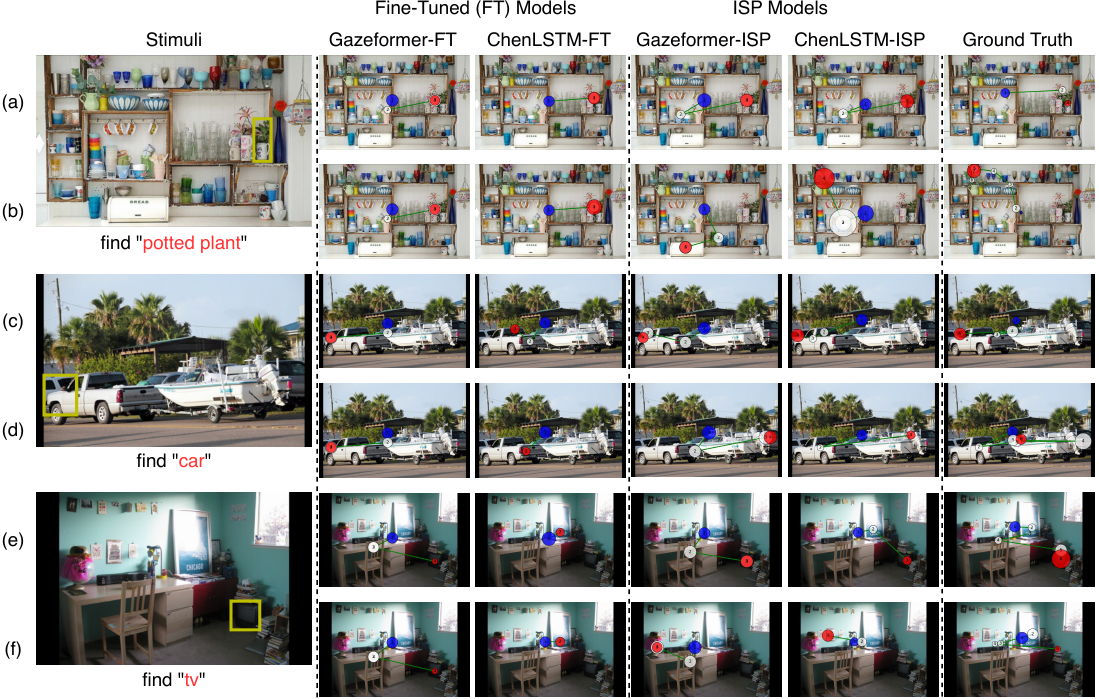}
\caption{Qualitative scanpath examples from Gazeformer-FT, ChenLSTM-FT, Gazeformer-ISP, ChenLSTM-ISP, and ground truth on COCO-Search18~\cite{zhibo:2020:cocosearch}. The first column shows the stimuli with the target object, and each row compares model predictions with one observer's ground truth. The gaze patterns reveal whether observers successfully find the target object. Some successfully find it, focusing on the (a) potted plant, (c) car, and (e) TV, while others fail because they misrecognize (b) flowers, (d) other vehicles, or (f) searching around the wrong place.
\textcolor{blue}{Blue} and \textcolor{red}{red} dots mark the start and end of scanpaths, while \textcolor{yellow}{yellow} boxes represent the search target.}
\label{fig:sup_cocosearch_qual}
\end{figure*}

\begin{figure*}[h]
\centering
\includegraphics[width=1.0\linewidth]{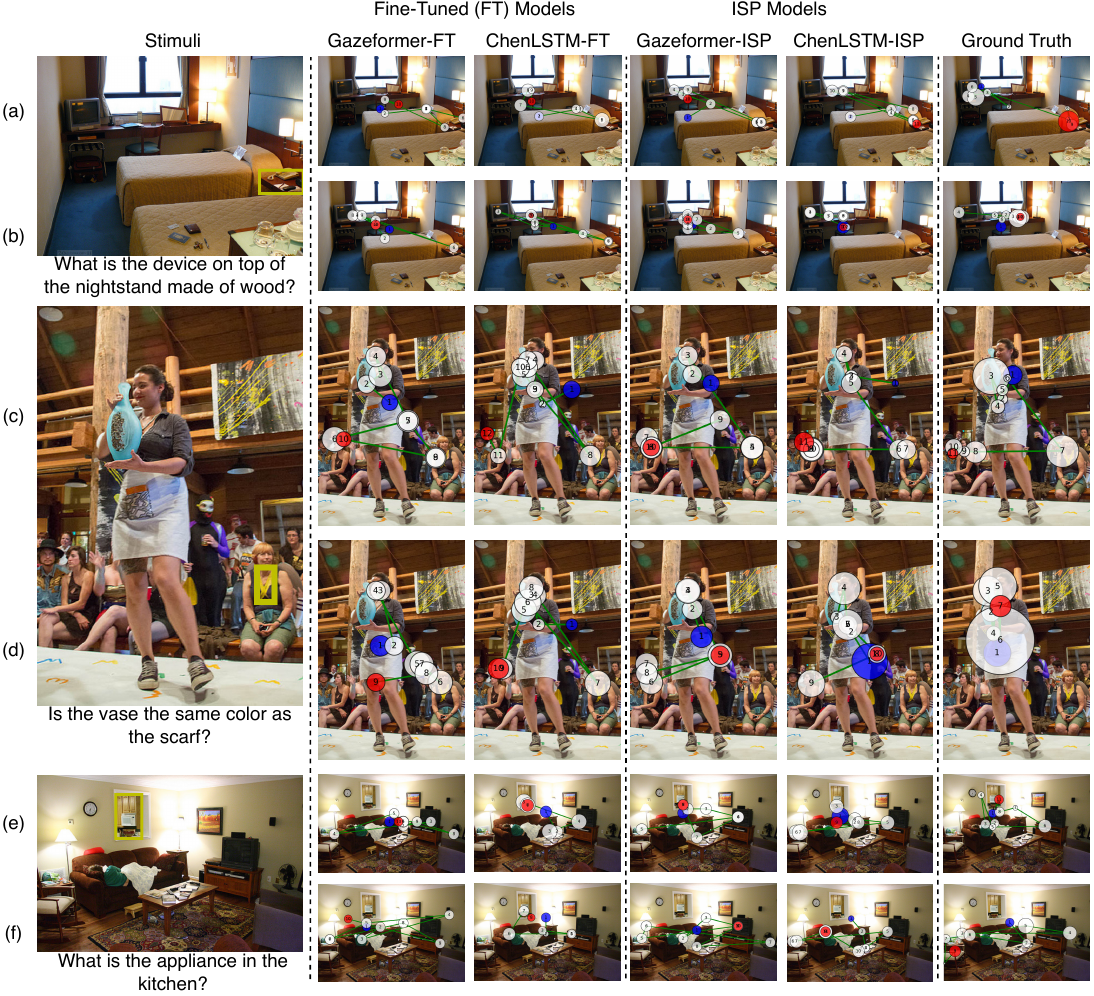}
\caption{
Qualitative scanpath examples from Gazeformer-FT, ChenLSTM-FT, Gazeformer-ISP, ChenLSTM-ISP, and ground truth on AiR-D~\cite{shi:2020:air}. The first column shows the stimuli with the corresponding question, and each row compares model predictions with one observer's ground truth. Examining gaze patterns reveals how correct answers correlate with observers focusing on specific regions like (a) the nightstand, (c) the scarf, and (e) the kitchen. In contrast, incorrect answers result from observers overlooking crucial objects: (b) searching the table instead of the nightstand, (d) not seeing the scarf, and (f) only exploring the living room instead of the kitchen. \textcolor{blue}{Blue} and \textcolor{red}{red} dots mark the start and end of scanpaths, while \textcolor{yellow}{yellow} boxes highlight important objects for answering correctly.}
\label{fig:sup_air_qual}
\end{figure*}

\textbf{OSIE-ASD.} 
Figure~\ref{fig:sup_autism_qual} illustrates a qualitative comparison on the OSIE-ASD~\cite{shuo:2015:austim} dataset. Examples (a), (c), and (e) show the scanpaths of observers with ASD, while examples (b), (d), and (f) show the scanpaths of controls. In examples (a) and (b), both Gazeformer-ISP and ChenLSTM-ISP effectively distinguish observers with ASD (repeatedly focusing on the center) from the control (exploring the dog and two children). Similarly, in panels (c) and (d), both models discern observers with ASD (repeatedly focusing on the center or the same object for an extended period) from the control (exploring the boys and ball). The same distinction is observed in panels (e) and (f). In contrast, fine-tuned baselines like Gazeformer-FT and ChenLSTM-FT lack this ability to discern such distinctions in gazing patterns across observers.

\textbf{COCO-Search18.} Figure~\ref{fig:sup_cocosearch_qual} presents a qualitative comparison on the COCO-Search18~\cite{zhibo:2020:cocosearch} dataset. In examples (a) and (b), observers search for the target `potted plant' with different patterns. Gazeformer-ISP and ChenLSTM-ISP reveal that observer (a) successfully finds the plant, while observer (b) fails to find it. Similarly, in examples (c) and (d), observers search for the target `car' with different patterns. Gazeformer-ISP and ChenLSTM-ISP show that observer (c) successfully finds the car, while observer (d) fails, by recognizing the vehicle on the right-hand side of the image as a car. Lastly, in examples (e) and (f), observers search for the target `TV' with different patterns. Gazeformer-ISP and ChenLSTM-ISP demonstrate that observer (e) successfully finds the TV, while observer (f) fails, fixating around the table. Fine-tuning methods Gazeformer-FT and ChenLSTM-FT do not clearly indicate success or failure in finding the target object.

\textbf{AiR-D.} Figure~\ref{fig:sup_air_qual} offers a qualitative comparison on the AiR-D~\cite{shi:2020:air} dataset. In examples (a) and (b), two observers respond differently to the question `What is the device on top of the nightstand made of wood?' Gazeformer-ISP and ChenLSTM-ISP reveal that observer (a) correctly identifies the answer `phone' by focusing on the nightstand, while observer (b) fails due to attention on the table, not the nightstand. Similarly, in examples (c) and (d), observers answer `Is the vase the same color as the scarf?' differently. Gazeformer-ISP and ChenLSTM-ISP show that observer (c) correctly answers `no' by examining colors, while observer (d) fails to find the scarf. Lastly, in examples (e) and (f), observers respond differently to `What is the appliance in the kitchen?' Gazeformer-ISP and ChenLSTM-ISP show observer (e) correctly finding the `microwave' in the kitchen, while observer (f) fails, focusing mostly on the living room. Fine-tuned baselines like Gazeformer-FT and ChenLSTM-FT cannot explain reasons for observers' correct or incorrect answers based on predicted scanpaths.

\end{document}